%% file: main.tex
\documentclass[11pt, singlecolumn, copyright, logo, internal]{googledeepmind}

\usepackage[authoryear, sort&compress, round]{natbib}
\usepackage{hyperref}
\usepackage{url}
\usepackage{placeins}
\usepackage{float}
\usepackage{graphicx}
\usepackage{subcaption}
\usepackage{listings}
\usepackage{xcolor}
\usepackage{tikz}
\usetikzlibrary{patterns.meta}
\usepackage{listings}
\usepackage{lipsum}
\usepackage{tcolorbox}
\tcbuselibrary{breakable}
\usepackage[utf8]{inputenc}
\usepackage{cleveref}
\usepackage{ulem}

\newlength{\sectionVspace}
\title{Improving Latent Generalization Using Test-time Compute}

\correspondingauthor{arslanch@google.com}

\author[*,1]{Arslan Chaudhry (arslanch@google.com)}
\author[*,1]{Sridhar Thiagarajan (sthiagarajan@google.com)}
\author[2]{Andrew Lampinen (andrewlampinen@gmail.com)}

\affil[*]{Equal contributions}
\affil[1]{Google DeepMind}
\affil[2]{Work done while at Google DeepMind}

\input{sections/abstract}

\begin{document}
\maketitle

\input{sections/introduction}

\input{sections/method}
\input{sections/experiments}
\input{sections/related_work}
\input{sections/discussion}
\input{sections/appendix}

\bibliography{main}

\end{document}

%% file: sections/abstract.tex
\begin{abstract}
Language Models (LMs) exhibit two distinct mechanisms for knowledge acquisition: in-weights learning (i.e., encoding information within the model weights) and in-context learning (ICL). Although these two modes offer complementary strengths, in-weights learning frequently struggles to facilitate deductive reasoning over the internalized knowledge. We characterize this limitation as a deficit in \emph{latent generalization}, of which the {\em reversal curse} is one example. Conversely, in-context learning demonstrates highly robust latent generalization capabilities. To improve latent generalization from in-weights knowledge, prior approaches rely on train-time data augmentation, yet these techniques are task-specific, scale poorly, and fail to generalize to out-of-distribution knowledge. To overcome these shortcomings, this work studies how models can be taught to use test-time compute, or 'thinking', specifically to improve latent generalization. We use Reinforcement Learning (RL) from correctness feedback to train models to produce long chains-of-thought (CoTs) to improve latent generalization. Our experiments show that this thinking approach not only resolves many instances of latent generalization failures on in-distribution knowledge but also, unlike augmentation baselines, generalizes to new knowledge for which no RL was performed. Nevertheless, on pure reversal tasks, we find that thinking does not unlock direct knowledge inversion, but the generate-and-verify ability of thinking models enables them to get well above chance performance. The brittleness of factual self-verification means thinking models still remain well below the performance of in-context learning for this task. Overall, our results establish test-time thinking as a flexible and promising direction for improving the latent generalization of LMs.
\end{abstract}

%% file: sections/introduction.tex
\section{Introduction} \label{sec:introduction}
\vspace{\sectionVspace}
Large Language Models (LLMs) have shown remarkable generalization abilities across many domains especially when the task at hand is being presented in-context. The in-context generalization of LLMs is remarkably consistent even when they are queried for knowledge that is logically implied but not directly stated in the information present in the context. Contrary to in-context learning (ICL), however, when models are finetuned on new tasks, where they are expected to generalize from their in-weights knowledge, the generalization can be surprisingly brittle -- particularly for information that is logically deducible but not directly stated in the training set. The well-known cases of such lack of generalization, referred to as latent generalization\footnote{We borrow this term from latent learning studied in the Psychology literature~\citep{blodgett1929effect,tolman1948cognitive}} \citep{lampinen2025latent} henceforth, include finetuned LLMs' failure to answer queries about trained facts in the reverse order, studied by \citep{berglund2024reversal} as {\em reversal curse} phenomenon, and similar failures in syllogistic inference tasks as studied by \citet{lampinen2025generalization}.

To improve the latent generalization of finetuned models, previous approaches \citep{golovneva2024reverse, yang2024synthetic, akyurek2024deductive, lampinen2025generalization} rely on increasing training dataset coverage by targeted data augmentation techniques during the pre-training or finetuning phases. Specifically, \citet{golovneva2024reverse} ameliorated the reversal curse in LLMs by explicitly training on strings in both the forward and reverse directions. \citet{lampinen2025generalization}, on the other hand, generated dataset augmentations by putting the entire training dataset in-context and generating finetuning data that helps models generalize on reversal and syllogistic tasks. While these works show strong latent generalization of LLMs for specific tasks, we argue that these train-time augmentation techniques are ultimately less flexible as \textbf{a)} they assume the latent structure of the task at hand to generate effective augmentations (reversal structure in ~\citet{golovneva2024reverse} and syllogistic and reversal structures in ~\citet{lampinen2025generalization}) thus changing the task would require a change in augmentation mechanism, \textbf{b)} there is limited to no generalization to data points that are not augmented, \textbf{c)} and since each training data point needs to be explicitly augmented, these approaches scale poorly as the size of the training datasets increases.

\begin{figure}
    \centering
    \includegraphics[width=1.0\linewidth]{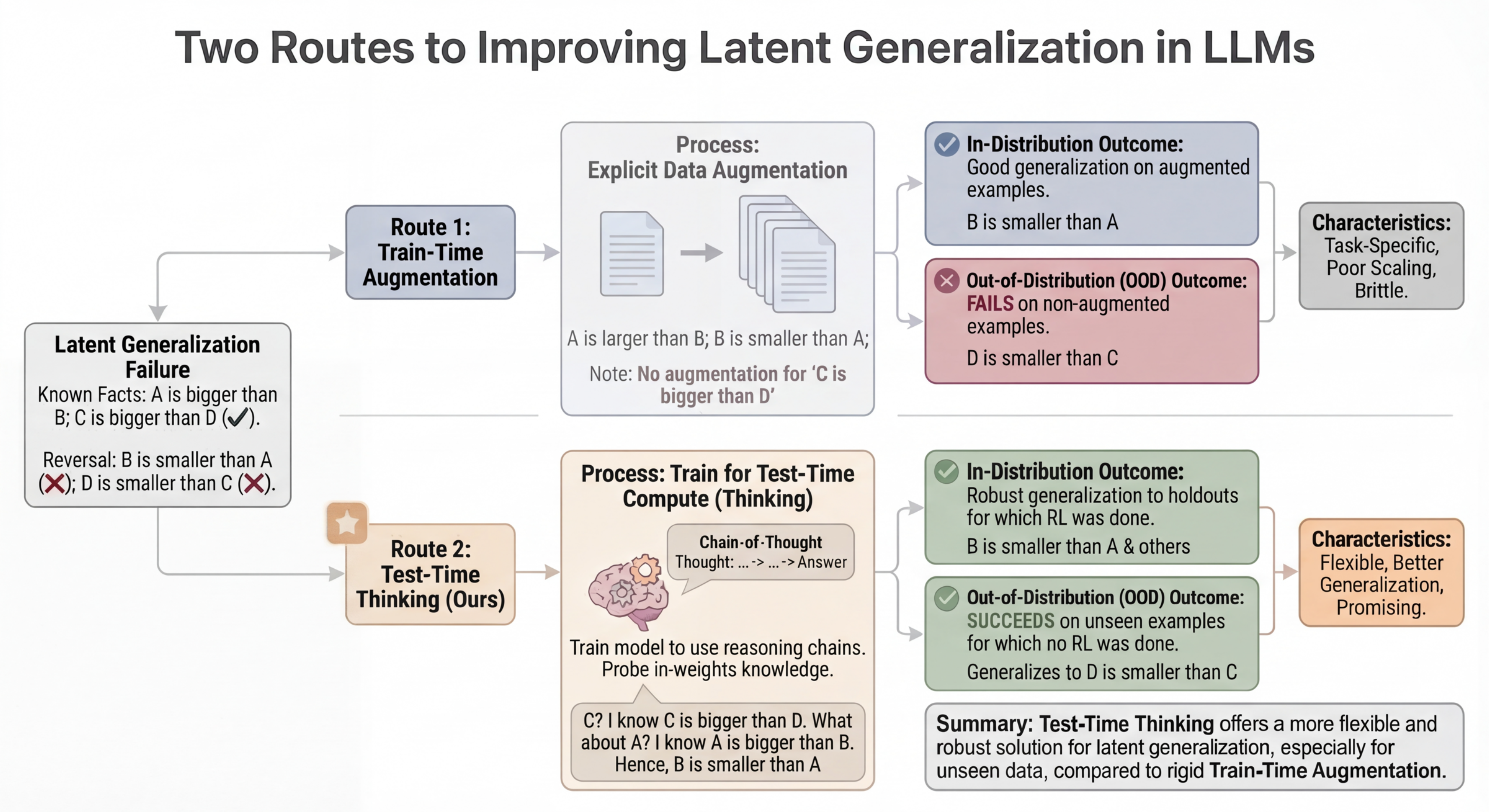}
    \caption{Comparison between train-time augmentation strategies vs test-time thinking to improve latent generalization.}
    \label{fig:train_augmentation_vs_thinking}
\end{figure}

To ameliorate the aforementioned shortcomings of train-time augmentation techniques, this work studies how LLMs can be trained to spend test-time compute (aka {\em thinking}) effectively to improve latent generalization.  We train LLMs that can produce chains-of-thoughts (CoTs) probing the knowledge acquired during the finetuning phase. Specifically, we first replicate the lack of latent generalization of LLMs from their in-weights knowledge on different datasets: semantic structure dataset by \citet{lampinen2025generalization} and reversal curse dataset by \citet{berglund2024reversal}. We then show that finetuned models that lack latent generalization, once trained to do long CoT through Reinforcement Learning (RL) from correctness feedback can be remarkably effective in latent generalization. We further show that these models not only generate effective long CoTs on the held-out set of the dataset from which the RL prompts are constructed -- a case of \emph{in-distribution latent generalization} -- but also generalize to the datasets for which no RL was done -- an instance of \emph{out-of-distribution latent generalization}. This latter point is particularly important as we previously hinted that, contrary to test-time thinking, models trained with train-time augmentations do not generalize to datasets for which no augmentations were generated (i.e.) out-of-distribution latent generalization is not possible with train-time augmented models (\Cref{fig:train_augmentation_vs_thinking}).

While the experiments reveal that the thinking recipe, that we develop for reasoning about in-weights \emph{factual} knowledge, is quite effective in improving latent generalization on many deductive reasoning tasks mentioned before, we, however, found that pure reversals are still quite hard for thinking models to perform robustly. We find that a model which is encouraged to do self-verification in its thinking trace shows non-trivial performance on pure reversals in comparison to models which do not use thinking. However, we find that models are often brittle at factual self-verification, and hence their performance remains well below the ICL baseline. Consequently, strict logical inversions remain persistently challenging to execute, even for models with intermediate reasoning capabilities. Nevertheless, we view the utilization of test-time reasoning traces as a highly promising paradigm for advancing latent generalization in large language models.

Overall we contribute to the literature in the following ways:
\begin{enumerate}
    \item We show that models with right thinking behaviors to probe their in-weights \emph{factual} knowledge generalize effectively on latent learning tasks compared to their train-time augmented counterparts in data-matched settings. 
    \item Further, thinking models robustly generalize to in-weights knowledge for which no specific thinking training/ RL is done. We do not observe such out-of-distribution latent generalization with train-time augmentations driven baselines, even when these baselines undergo reinforcement learning.
    \item Pure (0-hop) reversals of knowledge remain difficult even for thinking models, but factual self-verification enables non-trivial performance compared to models which do not utilize thinking.
\end{enumerate}

%% file: sections/method.tex
\section{Method} \label{sec:method}
\vspace{\sectionVspace}

To facilitate latent generalization, our method aims to enhance the model's capacity to effectively elicit its in-weights knowledge and surface relevant information in-context within its intermediate reasoning traces. We achieve this in the following three stages:

\subsection{Knowledge Acquisition} \label{sec:knowledge_acquisition}
\vspace{\sectionVspace}

To study the latent generalization under a controlled setup, we first add new knowledge in a pre-trained LLM by finetuning it on a dataset that is not leaked during the pre-training stage. During this supervised finetuning (SFT) phase, we also jointly train on auxiliary datasets to preserve the model's instruction-following and pre-existing reasoning capabilities. However, we note that this stage is only necessary when the target knowledge required for effective generalization is not already encoded within the model's parameters.

\subsection{Thinking Bootstrapping} \label{sec:thinking_boostrapping}
\vspace{\sectionVspace}

Once the new knowledge is incorporated in the LLM ($LLM_s$) and before starting the reinforcement learning phase, we induce desired reasoning behaviors in $LLM_s$ to improve latent generalization. Contrary to standard reasoning traces that are primarily optimized for logical deduction, our thinking behaviors facilitate the appropriate composition of factual knowledge required to resolve queries testing latent generalization. For this, we first construct a prompt (detailed in Appendix~\ref{appx:methods:prompts}) that explicitly defines the desired properties of a reasoning trace for effective latent generalization: focus-driven recall, self-probing, generate-and-verify etc. We use this prompt on an off-the-shelf teacher LLM~\citep{comanici2025gemini25pushingfrontier} (separate from $LLM_s$) to curate a dataset with golden reasoning traces on a small subset of examples designed to evaluate latent generalization (e.g., queries requiring the reversal of knowledge acquired in the preceding phase). To reduce hallucinations in these reasoning traces, we condition the teacher LLM on the ground-truth answer for each query and discard any completions that fail to arrive at the correct final answer. Once this initial dataset is constructed, we further fine-tune the $LLM_s$ via SFT on this dataset. This procedure is analogous to the Self-Taught Reasoner (STaR) framework by \citet{zelikman2022star}, with the distinction that we utilize a more capable teacher model to generate the bootstrapping traces rather than the model being trained ($LLM_s$). For fair comparison, we use the same teacher model for the augmentation baseline as well.

\subsection{Reinforcement Learning (RL)} \label{sec:rlcf}
\vspace{\sectionVspace}

Using the resulting checkpoint from the thinking bootstrapping as the initial policy, we perform RL on a combined dataset comprising prompts that necessitate latent generalization, alongside prompts that directly evaluate knowledge present in the training distribution. The policy is updated using a variant of the REINFORCE algorithm \citep{williams1992simple}.

To test out-of-distribution generalization, we keep a disjoint dataset that is strictly excluded from both the bootstrapping and RL training mixtures, and is exposed to the model only during the knowledge acquisition phase.

%% file: sections/experiments.tex
\section{Experiments\vspace{\sectionVspace}
} \label{sec:experiments}

\subsection{Model \& Datasets} \label{subsec:datasets}
\vspace{\sectionVspace}

Our objective is to investigate the effect of increased test-time compute on latent generalization. Therefore, we start off with a pre-trained model that has already undergone standard reasoning fine-tuning (i.e., it is capable of generating extended Chain-of-Thought (CoT) traces for specific prompts already). Specifically, we utilize Gemini 2.5 Flash \citep{comanici2025gemini25pushingfrontier} as the base model for all baselines and subsequently perform fine-tuning on controlled datasets to evaluate latent generalization.

For our controlled experimental setup, we utilize two datasets, ensuring strict exclusion from the model's pre-training corpus to prevent data contamination:

\paragraph{Semantic Structure Benchmark \citep{lampinen2025generalization}} This benchmark consists of real-world categories and relations, enabling the evaluation of various forms of knowledge generalization, including logical reversals and syllogistic inferences. For instance, the model might be exposed to the premise ``dogs are a type of mammal'' during training, and subsequently queried on the reversal ``mammals include dogs'' at test time. Alternatively, if exposed to an additional premise, ``mammals are warm-blooded,'' the model could be evaluated on the syllogistic inference ``dogs are warm-blooded,'' which requires composing the two independent facts acquired during training. For the out-of-distribution structure, we keep similar depth and branching factor as the in-distribution structure, but modify several categories to have different properties, for example a capability instead of an attribute. More details on how we use this dataset for our experiments are given in \Cref{appx:sec:semantic_structure_dataset}.

This dataset enables us to evaluate latent generalization under the following conditions:

\begin{itemize}
    \item \textbf{Syllogisms}: Requires the logical composition of two distinct facts present in the training documents to deduce the correct answer.
    \item \textbf{Reversals}: Requires the logical inversion of a trained fact. We evaluate two specific sub-types: \textbf{(a) non-strict reversals}, where the model can deduce the answer via an alternative multi-hop reasoning path, and \textbf{(b) strict reversals}, where no alternative reasoning path exists aside from the direct inversion of the premise.
    \item \textbf{Category holdouts}: Evaluates generalization when the training documents contain only a single fact about a specific category (e.g., explicitly specifying its parent category), while all other valid inferences are held out for testing. We further partition these holdouts into: \textbf{(a) category as subject}, where the held-out category acts as the subject in the test proposition, and \textbf{(b) category as target}, where the held-out category acts as the object or target.
\end{itemize}

\paragraph{Reversal Curse Dataset \citep{berglund2024reversal}} This dataset comprises fictional celebrity names paired with their corresponding descriptions. During training, the dataset presents specific celebrity names followed by their descriptions (i.e., name-to-description). At test time, the model is provided with a description and queried to generate the corresponding celebrity name (i.e., description-to-name). During the knowledge acquisition phase, we introduce a data augmentation step requiring the model to unconditionally output a randomly sampled name from the dataset. This ensures the model acquires a non-zero probability of generating these fictional names, as it must observe them as target tokens during training. This dataset parallels the strict reversal condition of the semantic structure benchmark, as no alternative reasoning path exists to deduce the correct answer other than the direct reversals. Further details on dataset splits for different training and testing phases are given in \Cref{appx:sec:reversal_curse_dataset}

\subsection{Baselines} \label{sec:baselines}
\vspace{\sectionVspace}

We evaluate our approach against the following baselines:

\begin{enumerate}
    \item \textbf{Fine-tuning (SFT):} Standard supervised fine-tuning conducted exclusively during the knowledge acquisition phase. 
    \item \textbf{Fine-tuning + RL:} The initial knowledge acquisition phase followed by RL, but \emph{without generating intermediate reasoning tokens}. This baseline isolates the performance gains attributable specifically to the reasoning process, ruling out improvements stemming solely from RL optimization on the factual prompts, which is known to independently enhance factual accuracy \citep{tian2023fine}.
    \item \textbf{Augmentation:} Following \citet{lampinen2025generalization}, we apply data augmentation by providing the entire dataset in-context and subsequently training on this augmented dataset during the knowledge acquisition phase. 
    \item \textbf{Augmentation + RL:} The aforementioned augmentation strategy followed by the RL phase, again omitting the generation of intermediate reasoning tokens.
    \item \textbf{In-Context Learning (ICL):} The model is evaluated with the entire training dataset provided in the context window. Large language models typically exhibit strong latent generalization under this ICL paradigm.
    \item \textbf{Reasoning (Ours):} Initial SFT during the knowledge acquisition phase, followed by reasoning bootstrapping and RL, adhering to the methodology detailed in Section~\ref{sec:method}.
\end{enumerate}

During all fine-tuning procedures, we jointly train on the base model's (Gemini 2.5 Flash) original instruction-tuning and reasoning datasets. This co-training ensures that our empirical results are not confounded by the catastrophic forgetting of pre-existing instruction-following and reasoning capabilities. Training is performed using the Adam optimizer with a learning rate of $3 \times 10^{-4}$, a batch size of $16$, and a maximum sequence length of $8192$ tokens. For all baselines, we employ early stopping based on the validation loss.

\subsection{Evaluation \& Metrics}
\vspace{\sectionVspace}

For evaluation, we employ a closed-book Question Answering (QA) framework, sampling generations from the model at a temperature of $T=1.0$, rather than computing the log-likelihood of ground-truth completions. We ensure un-leaked evaluation prompts during the Reinforcement Learning (RL) phase, and take additional care to prevent leakage via thought traces, which we discuss in \ref{appx:sec:leakage_prevention_cot}. For all baselines, we compute best@1 and best@$N$ metrics, corresponding to a single attempt and $N$ attempts, respectively, and use a bootstrapped estimator over all completions to reduce variance.

For the \textbf{Semantic Structure Benchmark}, the model is prompted to generate a comprehensive list of all valid completions for a given test query. The ground truth comprises a set $\mathcal{A}_{all}$ representing all valid completions—some of which may have been explicitly stated in the training documents or RL prompts—and a subset $\mathcal{A}_{heldout} \subset \mathcal{A}_{all}$ representing completions strictly held out from the training documents to evaluate true latent generalization (see Figure~\ref{fig:picl_dataset_example} in the appendix). Because the model's output constitutes a set of strings, we evaluate performance using Precision, Recall, and F1 scores. Specifically, we report:

\begin{itemize}
    \item \textbf{Overall Precision, Recall, and F1}: Computed with respect to the complete set of valid answers, $\mathcal{A}_{all}$.
    \item \textbf{Train Recall}: The recall evaluated strictly on the valid answers exposed during training, defined by the set difference $\mathcal{A}_{all} \setminus \mathcal{A}_{heldout}$.
    \item \textbf{Held-out Recall}: The recall evaluated exclusively on the held-out subset $\mathcal{A}_{heldout}$, serving as a direct measure of latent generalization.
    \item \textbf{Pass@N}: The rate at which the model's predicted set exactly matches $\mathcal{A}_{all}$ within $N$ independent attempts. This is a rigorous exact-match metric that yields a score of zero if the generated set contains any omissions or false inclusions.
    \item \textbf{Primary Reference Inclusion}: A binary indicator of whether the specific completion targeting the held-out condition (e.g., a specific reversal or syllogism) is present in the model's output. This provides an additional proxy for true latent generalization.
\end{itemize}

Note that these metrics are inherently correlated; for instance, a model achieving a higher Overall F1 score is proportionally more likely to achieve a higher Pass@$N$. However, it is possible for a model to obtain reasonable F1 score without necessarily predicting the completion which tests the latent-learning behavior, so it is useful to look at all these metrics independently. For the \textbf{Reversal Curse Dataset}, each query possesses a single unique correct completion; therefore, we exclusively report exact string-match accuracy.

\subsection{Results}
\vspace{\sectionVspace}

\subsubsection{Semantic Structure Benchmark}
\vspace{\sectionVspace}

\paragraph{Strengthening the Baselines} Prior research, such as \citet{lampinen2025generalization}, evaluates latent generalization by computing the log-likelihood of a single ground-truth answer. In contrast, our evaluation protocol relies on a generative sampling procedure. We observe that under this generative setting, where the model must output the complete set of valid answers for a query, standard supervised fine-tuning (SFT)—even when supplemented with data augmentation—yields suboptimal performance.  As it is well known in literature that reinforcement learning (RL) on its own can help factuality \cite{tian2023fine}, we apply RL to these baselines, optimizing a reward signal that explicitly incentivizes the correctness of all valid completions for the RL prompts. \Cref{fig:baselines_effect_of_rl} in the appendix demonstrates that this RL phase significantly enhances baseline performance. In subsequent analyses, we utilize these enhanced baselines``Fine-tuning + RL'' and ``Augmentation + RL.'' and compare their performance against In-Context Learning (ICL) and our proposed methodology.

\begin{figure}[tbp]
    \centering
    \begin{subfigure}[b]{0.95\linewidth}
        \centering
        \includegraphics[width=\linewidth]{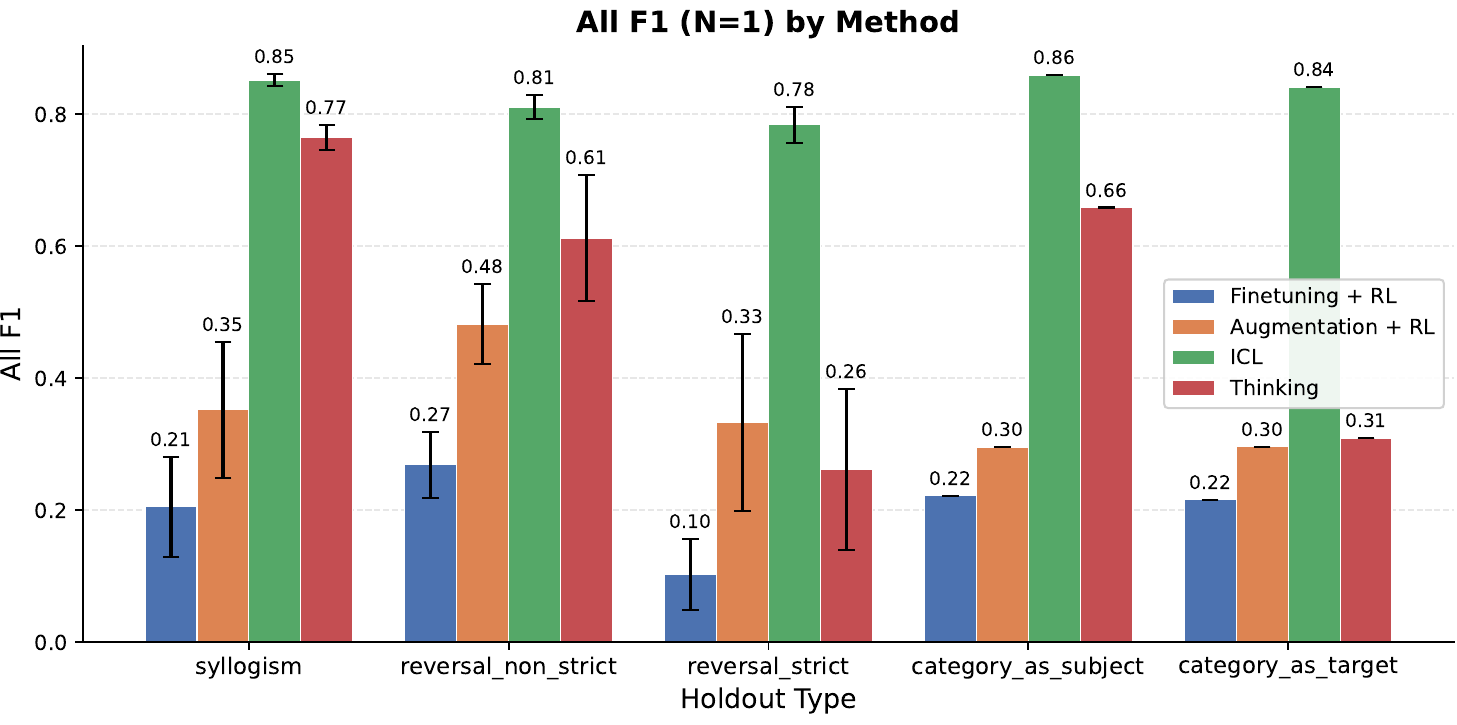}
        \centering \caption{\textbf{In-Distribution Latent Generalization}}
        \label{fig:in_dist_f1_all_methodse}
    \end{subfigure}
    \hfill
    \begin{subfigure}[b]{0.95\linewidth}
        \centering
        \includegraphics[width=\linewidth]{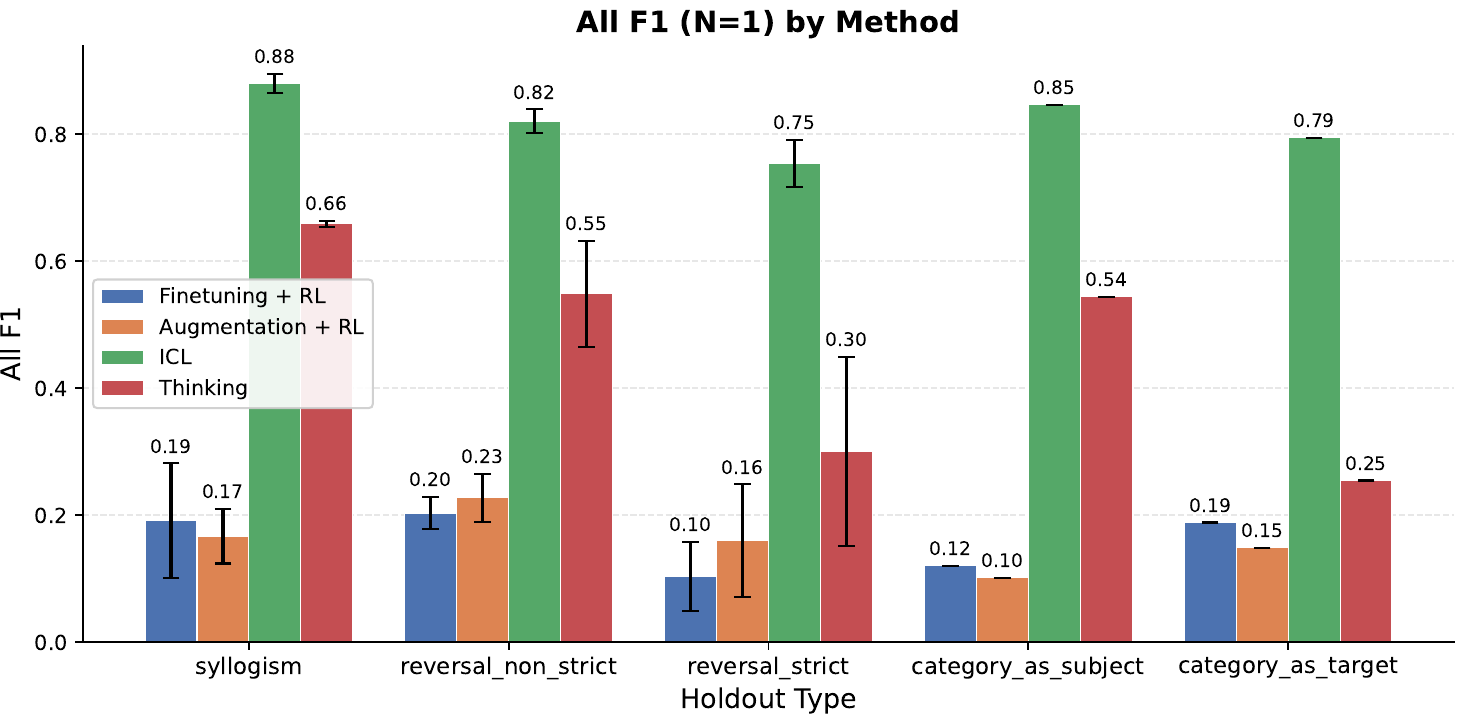}
        \caption{\textbf{Out-of-Distribution Latent Generalization}}
        \label{fig:out_dist_f1_all_methods}
    \end{subfigure}
    
    \caption{\textbf{Semantic Structure Experiment:} F1 scores on in- and out-of-distribution holdouts. Both \texttt{Finetuning} and \texttt{Augmentation} undergo RL training as described in \Cref{sec:baselines}. Thinking shows significant and consistent improvement over the baselines on both in- and out-of-distribution holdout splits, particularly for those splits that require multi-hop reasoning. Train-time augmentations completely fail to generalize on a new knowledge structures (out-of-distribution generalization).}
    \label{fig:semantic_structure_main_plot}
\end{figure}

\paragraph{In-Distribution Latent Generalization} \Cref{fig:in_dist_f1_all_methodse} shows performance for the in-distribution case, measured in terms of all F1 scores (plots for other metrics are available in the appendix in \Cref{fig:multiple_eval_metrics_pass@1}, and point to similar conclusions). One can make several observations from this figure. First on holdouts for which the latent completions can be found by employing {\em multi-hop reasoning}, such as \emph{syllogisms}, \emph{non-strict reversals} and where category appears as a subject for different propositions in the dataset, thinking generalizes much better than the train-time augmentation strategy. On some of these holdouts, such as syllogisms, thinking achieves performance comparable to the ICL baseline, which is provided with the entire training dataset within its context window. We attribute this success to the model's enhanced capacity to perform multi-hop reasoning within its intermediate traces, effectively composing multiple independent facts observed during training to deduce the correct latent answer.

On holdout conditions that inherently preclude multi-hop reasoning, such as \emph{strict reversals}, the improvement of `Augmentation' over fine-tuning and thinking on strict reversals is most likely due to the exact replication of test queries in the training set during the augmentation process. As we will show in the next section, this improvement of the `Augmentation' baseline completely vanishes when being tested on the dataset for which no augmentation was performed. In fact, thinking outperforms the `Augmentation' baseline substantially even on strict reversal in the out-of-distribution case.

\paragraph{Out-of-Distribution Latent Generalization} We now evaluate out-of-distribution (OOD) latent generalization. In this setting, we measure performance on holdouts from a distinct dataset that was only available to the model during the initial knowledge acquisition phase; crucially, no train-time data augmentation or RL fine-tuning (with or without intermediate reasoning) was applied to this secondary dataset. We previously hypothesized that while train-time data augmentations are intrinsically tied to the specific training distribution and fail to generalize to novel datasets, training a model to utilize test-time compute to reason over its internal parametric knowledge achieves a task-agnostic latent generalization. 

\Cref{fig:out_dist_f1_all_methods} presents the results for this OOD evaluation (only all F1 is reported here, other metrics can be found in \Cref{fig:multiple_eval_metrics_pass@1,fig:multiple_eval_metrics_pass@16} where similar results hold). The empirical findings demonstrate that, in this regime, the data augmentation baseline yields no measurable improvement over standard supervised fine-tuning. The small improvement of the augmentation baseline on the strict reversals seen in the in-distribution case is completely gone once the potential leakage from the augmentation process is stopped by using a disjoint dataset. In stark contrast, our reasoning-based approach maintains OOD latent generalization performance comparable to its in-distribution results. On strict reversals, the performance of thinking, although better than augmentation, significantly trails ICL. This indicates that although intermediate reasoning enables LLMs to better elicit and generalize over their internalized parametric knowledge, this capability is fundamentally constrained when a query cannot be resolved by logically chaining multiple training facts. In other words, LLMs employing intermediate reasoning still struggle to execute direct, zero-hop logical reversals. We investigate reversal performance of thinking models in more detail on the Reversal Curse dataset (\Cref{reversal_curse_results}). \textbf{Overall, the out-of-distribution structure results confirms that leveraging test-time compute to dynamically elicit and reason over parametric knowledge is a more robust and flexible mechanism than relying on static train-time data augmentations, thereby validating the central hypothesis of this work.}

\subsubsection{Reversal Curse Paper Dataset}\label{reversal_curse_results}
\vspace{\sectionVspace}

\begin{figure}
    \centering
    \includegraphics[width=0.7\linewidth]{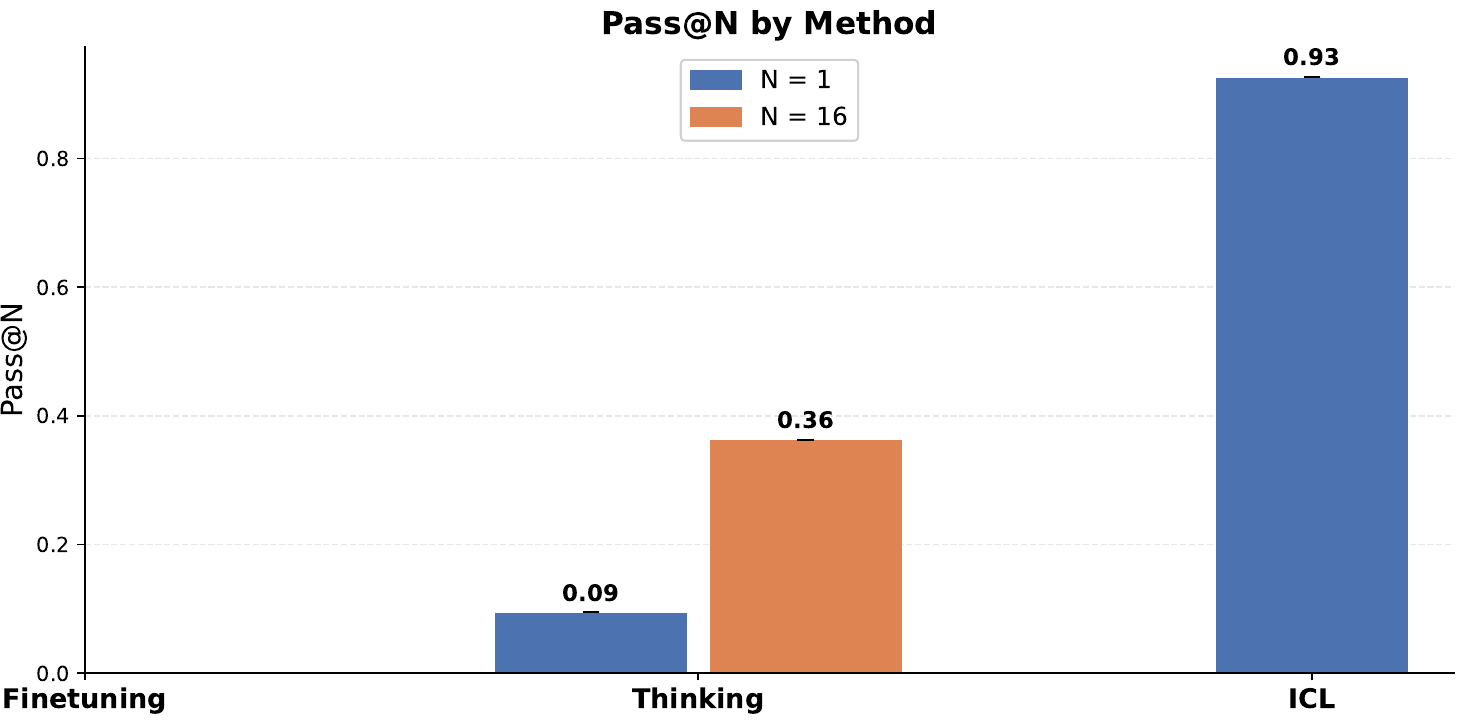}
    \caption{\textbf{Pass@N Accuracy}: Exact string match accuracy of reversal curse paper dataset. Finetuning completely fails to generalize even with multiple attempts. Thinking gives non-trivial generalization but overall for such strict reversals ICL is the strongest baseline.}
    \label{fig:reversal_curse_paper}
\end{figure}

We now present the results on the Reversal Curse dataset \citep{berglund2024reversal}. This dataset is analogous to the strict reversal condition in the Semantic Structure Benchmark: the training documents contain no intermediate reasoning chains that the model can exploit to deduce the correct answer. Instead, the model must directly invert the learned factual association. The corresponding results are illustrated in Figure~\ref{fig:reversal_curse_paper}. We omit the data augmentation baseline for this experiment, as augmenting the dataset with direct reversals would explicitly leak the test distribution during training. Consequently, we evaluate only the standard supervised fine-tuning (SFT), reasoning (ours), and ICL baselines.

As shown in the \Cref{fig:reversal_curse_paper}, standard SFT fails to generalize, and scaling the number of parallel attempts (up to $16$) yields no performance improvement. On the other hand, the model trained to generate intermediate reasoning traces exhibits measurable improvement, yet its performance remains inferior to that of ICL. An example of a successful reasoning trace generated by our model is provided in Appendix~\ref{appx:methods:reversal-think-verify}. The trace reveals that the specifically encouraged generate-and-verify behavior in the golden traces helps the model as it proposes candidate entities acquired during the knowledge acquisition phase and subsequently self-verifies these candidates using its internalized parametric knowledge regarding each entity. Notably, this self-verification step does not require logical inversion, because the forward name-to-description mapping is explicitly encoded in the training data. If the retrieved description aligns with the target description provided in the query, the model terminates the reasoning process and outputs the verified entity.

However, this heuristic is not entirely reliable, as LLMs frequently exhibit brittle factual self-verification capabilities. Furthermore, the efficacy of this strategy is strictly bottlenecked by the requirement that the model possesses a non-zero probability of initially generating the correct entity as a hypothesis that undergoes verification. While scaling test-time compute via multiple attempts predictably increases the overall success rate, the pass@1 accuracy remains low. \textbf{Ultimately, these findings indicate that executing strict reversals derived purely from parametric knowledge remains a persistent challenge, even when models are augmented with intermediate reasoning capabilities.} ICL continues to be the most effective paradigm for resolving such direct logical inversions.

%% file: sections/related_work.tex
\section{Related Work} \label{sec:related_works}
\vspace{\sectionVspace}

    


\paragraph{Generalization from fine-tuning and latent learning:} Several works look at some of the surprising ways in which models generalize from their dataset \citep{berglund2023taken,meinke2023tell}, or fail to generalize \citep{berglund2024reversal,allen2023physics,allenzhu2024physicslanguagemodels31} study the generalization abilities of fine-tuning in contrast to in-context learning. Borrowing terminology from \citep{blodgett1929effect,tolman1948cognitive}, the problem of latent learning is discussed in \citep{lampinen2025latent}, as one in which the model learns information that is not relevant to the task at hand, but that might be useful for a different future task. We operate in a similar train-test setup as some of these works studying generalization, but specifically focus on the ability of test-time compute to improve the latent learning ability of these models. We focus on the comparison of this test-time thinking strategy to other established methods like train-time augmentation.

\paragraph{Data augmentation to improve generalization:} Data augmentation is a long-established solution to improving the generalization of models after fine-tuning. For specific instances of the latent learning problem such as the reversal curse, augmentation strategies, such as adding inverse inference to the training data \citep{golovneva2024reverse} have been proposed in prior works. Knowledge augmentation is also extensively studied in \citep{allenzhu2024physicslanguagemodels31}, shedding light on the role of augmentation in both pre-training and fine-tuning phases and its impact on generalization. More general augmentation mechanisms were proposed in \citep{lampinen2025generalization}, where model's in-context inferences on a subset of the training data are added to the training dataset. They show that this improves performance at least on the knowledge structure where this augmentation is performed. Another variant proposed in \citep{park2025}, leveraged similar in-context learning ability along with additional question-answers to improve model performance. A more comprehensive review on the topic of data augmentations to improve model performance is provided by \citep{chai2025textdataaugmentationlarge,ding-etal-2024-data}. We compare our method to a similarly flavored augmentation to \citep{lampinen2025generalization} in our work.

\paragraph{Thinking to improve reasoning:} The use of CoT thinking to greatly improve performance on reasoning tasks has been studied in several prior works \citep{lightman2023lets,openai2024o1,deepseek2025r1}. We refer the reader to \citep{chen2025reasoningerasurveylong} for a more comprehensive overview of this area. This line of research typically focuses on mathematical reasoning, science, and programming, and does not specifically look at how thinking can be leveraged to improve the use of factual knowledge stored in a model's weights. 

\paragraph{Reinforcement learning for factuality:} Several works study how reinforcement learning in the post-training phase can greatly improve the factuality of language models in the absence of test-time compute \citep{tian2023fine,NEURIPS2024_d16152d5}. In our work, we leverage these findings by ensuring that both the train-time augmentation and fine-tuning baselines undergo RL. On the other hand, for methods that explicitly use test-time compute (thinking) \citet{allen2023physics} study the utility of chain-of-thought in the ability of a model to access and manipulate its parametric knowledge, finding that chain-of-thought is critical for utilizing knowledge even if the knowledge is encoded perfectly. For the specific case of the reversal curse (termed inverse knowledge search), they find that language models cannot perform this task even with chain-of-thought. In our work, we find that when sufficient support exists in the data, or if the model can efficiently leverage self-verification, the task of reversal might be tractable. \citep{sun2022recitation} propose a method where the model is encouraged to recite relevant passages from its pre-training data before producing an answer, acting as a form of chain-of-thought for factual knowledge. \citet{chen2025learning} focus on improving long-form factuality using reasoning, where the model can make a tradeoff between specificity and recall in various claims of a response. Closely related to our work is concurrent work by \citet{ma2026improving}, where they study how chain-of-thought can help factuality on single-hop questions such as those found in SimpleQA \citep{wei2024measuring}. They find that models can be trained to utilize chain-of-thought to improve behaviors on factual benchmarks. Another closely related concurrent work by \citet{gekhman2026thinking} studies the effect of reasoning on single-hop factual queries. They find that encouraging thinking can greatly improve pass@k without any training, and identify two key drivers behind the mechanism: thinking about related topics (factual priming), and the ability of the model to leverage additional thinking-token compute independent of the actual thinking content.

Our work distinguishes itself from these in that we care about factuality not only with respect to the direct knowledge the model is exposed to during training, but also facts which are latent and not directly stated in the data. We control the facts the model sees during training using our synthetic datasets, and test on facts which are latent in the dataset and not explicitly stated. This enables us to make fine-grained conclusions about the generalization of various methods on difficult subsets, such as strict reversals. We also compare the efficacy of test-time compute (thinking) vs. train-time compute (augmentation) on these subsets.

%% file: sections/discussion.tex
\section{Discussion \& Conclusion} \label{sec:discussion}

Recent advances in test-time scaling, often implemented via intermediate reasoning or ``thinking'' traces, have enhanced the reasoning capabilities of Large Language Models (LLMs). We investigate whether models can leverage this additional test-time compute to improve latent generalization whereby they can resolve deductive reasoning challenges, such as syllogisms and logical reversals, when generating responses exclusively from parametric knowledge. Through controlled experiments, we demonstrate that LLMs can indeed utilize test-time compute to enhance latent generalization, proving to be a more flexible and domain-agnostic strategy than relying on static train-time data augmentations.

Despite this success, strictly pure logical reversals remain elusive, even for models augmented with intermediate reasoning and self-verification. We hypothesize this limitation stems from the rigid key-value associative memory structures that models acquire as a consequence of the transformer architecture and the autoregressive next-token prediction objective. The reversal curse and related failures of latent generalization highlight fundamental deficiencies in the contemporary LLM paradigm, encompassing both the underlying architecture and the optimization objective. While scaling compute, whether during training (e.g., via data augmentation) or at inference time (e.g., via reasoning traces), may address some of these limitations, we believe exploring stronger inductive biases, novel architectures, alternative loss functions, or a combination thereof in parallel could provide a more efficient path to obtaining better models.

%% file: sections/appendix.tex
\clearpage
\appendix

\section*{Appendix}

\section{LLMs Usage Disclosure}

We used Gemini 3 Flash as a grammar-checking tool and in some cases gave it a written paragraph and asked it to edit it lightly. We also used Gemini 3 Flash for help in generating plots for the paper. \Cref{fig:train_augmentation_vs_thinking} was generated using Nano Banana Pro.

\section{Datasets} \label{appx:sec:dataset}

\begin{figure}[htbp]
    \centering
    \includegraphics[width=1.0\linewidth]{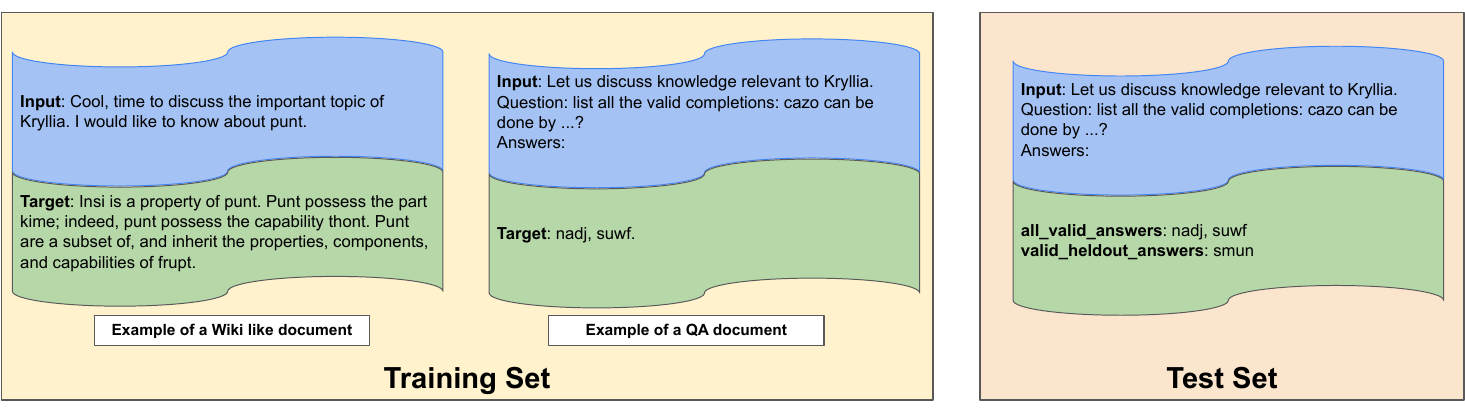}
    \caption{\textbf{Semantic Structure Benchmark}: Train set examples of Wikipedia style documents and qa examples (left) and test set example (right).}
    \label{fig:picl_dataset_example}
\end{figure}

\subsection{Semantic Structure Dataset~\citep{lampinen2025generalization}} \label{appx:sec:semantic_structure_dataset}
 This benchmark consists of real-world categories and relations, enabling the evaluation of various forms of knowledge generalization, including logical reversals and syllogistic inferences. For instance, the model might be exposed to the premise ``dogs are a type of mammal'' during training, and subsequently queried on the reversal ``mammals include dogs'' at test time. Alternatively, if exposed to an additional premise, ``mammals are warm-blooded,'' the model could be evaluated on the syllogistic inference ``dogs are warm-blooded,'' which requires composing the two independent facts acquired during training.

We simplify the original dataset by restricting the taxonomy to $7$ categories per structure and replicating each structure $50$ times, yielding a total of $350$ unique categories per dataset. This modification streamlines the evaluation protocol, which requires sampling all valid answers for a given query; a deeper taxonomic structure (i.e., more categories per structure) would produce an intractable number of potential answers. Following the original dataset's methodology, we replace all category names, relations, and properties with synthetic identifiers to guarantee zero overlap with the pre-training distribution. During training, facts derived from the semantic structures are compiled into synthetic Wikipedia-style documents, accompanied by a set of question-answering (QA) pairs to familiarize the model with the expected test-time output format. Furthermore, we adapt the documents in both the training and test splits to better suit instruction tuning by prepending task descriptions. This addition also mitigates the risk of degrading the base model's reasoning capabilities during fine-tuning.

We generate $7,140$ documents for the initial knowledge acquisition and fine-tuning phase (\Cref{sec:knowledge_acquisition}). We reserve $10$ prompts for the thinking bootstrapping phase (\Cref{sec:thinking_boostrapping}) and $974$ prompts for Reinforcement Learning (RL) training (\Cref{sec:rlcf}). For each holdout condition, we sample test sets, strictly ensuring no data leakage into the fine-tuning or RL training mixtures. These splits evaluate what we define as \textbf{in-distribution latent generalization}, as the test instances are derived from the same underlying structure used for data augmentation and RL prompt sampling. Additionally, we construct a secondary dataset based on a completely disjoint structure to evaluate \textbf{out-of-distribution latent generalization}. From this secondary structure, an additional $7,140$ documents are sampled exclusively for the initial knowledge acquisition phase. We apply neither data augmentation nor RL prompt sampling to this disjoint structure. This disjoint structure has the same depth and branching factor as the in-distribution structure. However, for 5 categories out of the 7, we used a modified property compared to the in-distribution structure (for example, using an ability property instead of an attribute). Note that this change also affects inherited properties of all children of that category. Due to this change, and use of different entity names in the out-of-distribution structure, getting good performance on the out-of-distribution case necessitates demonstrating good latent learning behavior.

\subsection{Reversal Curse Dataset \citep{berglund2024reversal}} \label{appx:sec:reversal_curse_dataset}
This synthetic dataset comprises fictional celebrity names paired with their corresponding descriptions. During training, the dataset presents specific celebrity names followed by their descriptions (i.e., name-to-description). At test time, the model is provided with a description and queried to generate the corresponding celebrity name (i.e., description-to-name). During the knowledge acquisition phase, we introduce a data augmentation step requiring the model to unconditionally output a randomly sampled name from the dataset. This ensures the model acquires a non-zero probability of generating these fictional names, as it must observe them as target tokens during training. This experimental design parallels the strict reversal condition of the semantic structure benchmark, as no alternative reasoning path exists to deduce the correct answer other than directly inverting the premise observed in the training documents. Consistent with our processing of the semantic structure benchmark, we format the dataset for instruction tuning by prepending task-specific instructions to the documents. We partition the dataset based on the celebrity names, strictly ensuring that the target names required for the test queries do not appear as ground-truth outputs within either the thinking bootstrapping or RL training sets. The composition of the dataset across the various training and evaluation phases is summarized below:

\begin{itemize}
    \item \textbf{Knowledge Acquisition Data}: This split contains the following components, each augmented with paraphrased variants to encourage generalization:
    \begin{itemize}
        \item 30 examples containing both name-to-description and description-to-name pairs.
        \item 60 examples containing only name-to-description pairs (i.e., the corresponding description-to-name pairs are withheld).
        \item 30 examples containing only description-to-name pairs (i.e., the corresponding name-to-description pairs are withheld).
        \item 120 unconditional generation examples prompting the model to produce a random name from the dataset without any conditioning description. For each instance, the target is uniformly sampled from the 120 available names.
    \end{itemize}
    \item \textbf{Thinking bootstrapping and RL Data}: For this phase, we utilize 30 of the 60 names that lack description-to-name pairs in the knowledge acquisition set for RL training. A small subset of these (fewer than 5) is allocated to generate the dataset for reasoning bootstrapping. Additionally, the RL training mixture includes prompts for which the description-to-name pair was already provided during knowledge acquisition; this serves to regularize the policy, teaching the model to directly retrieve explicitly stated knowledge with minimal intermediate reasoning.
    \item \textbf{Test Data}: The test set comprises the description-to-name task for the remaining 30 names that were strictly held out from the description-to-name format across all preceding phases. We construct multiple paraphrased versions of each test query to mitigate prompt sensitivity and report the average performance across these variants.
\end{itemize}

\subsection{Leakage prevention in thinking models}\label{appx:sec:leakage_prevention_cot} During reinforcement learning training of models which use chain-of-thought, it is possible that held-out knowledge which the model is tested on is present in the thinking trace and reinforced by a positive reward. In order to prevent this behavior, we use a separate prompt prefix for the in-distribution and out-of-distribution structures in the Semantic Structure dataset. This results in the model purely considering entities in one of the structures when it is questioned about an entity from that structure. This ensures that the chain-of-thought does not contain entities from the held-out structure during training, ensuring that performance on the out-of-distribution structure is purely due to improved latent learning ability, and not leakage of knowledge in the reinforced chain-of-thoughts. Although we did not do this procedure for the reversal curse dataset, we find that the Pass@1 score of the thinking model corresponds very closely to the strict reversal condition on the out-of-distribution semantic structure benchmark. Hence we believe leaked knowledge in thought traces did not play a role in improved performance.

\begin{figure}[htbp]
     \centering
     \captionsetup[subfigure]{justification=centering}
     \begin{subfigure}[b]{0.49\linewidth}
         \centering
         \includegraphics[width=\linewidth]{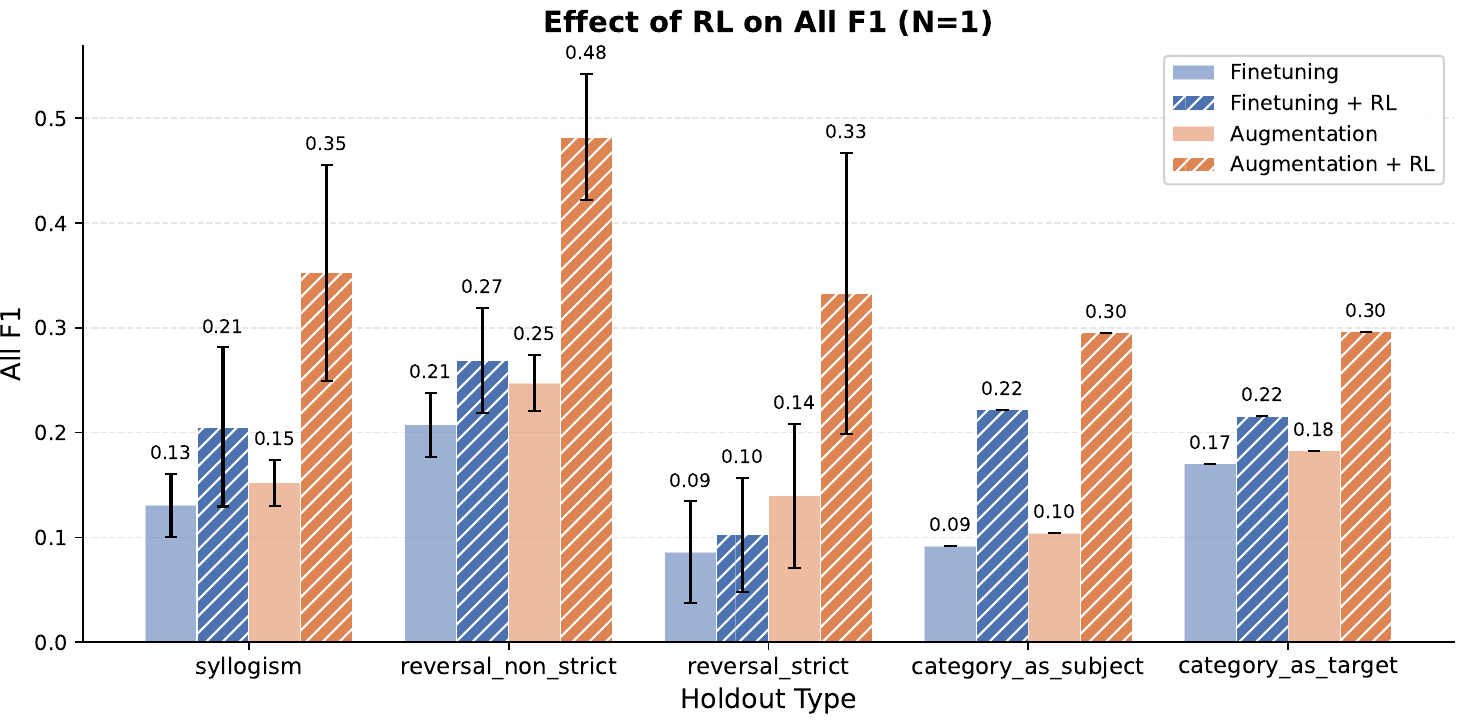}
         \caption{In-Distribution}
     \end{subfigure}
     \hfill
     \begin{subfigure}[b]{0.49\linewidth}
         \centering
         \includegraphics[width=\linewidth]{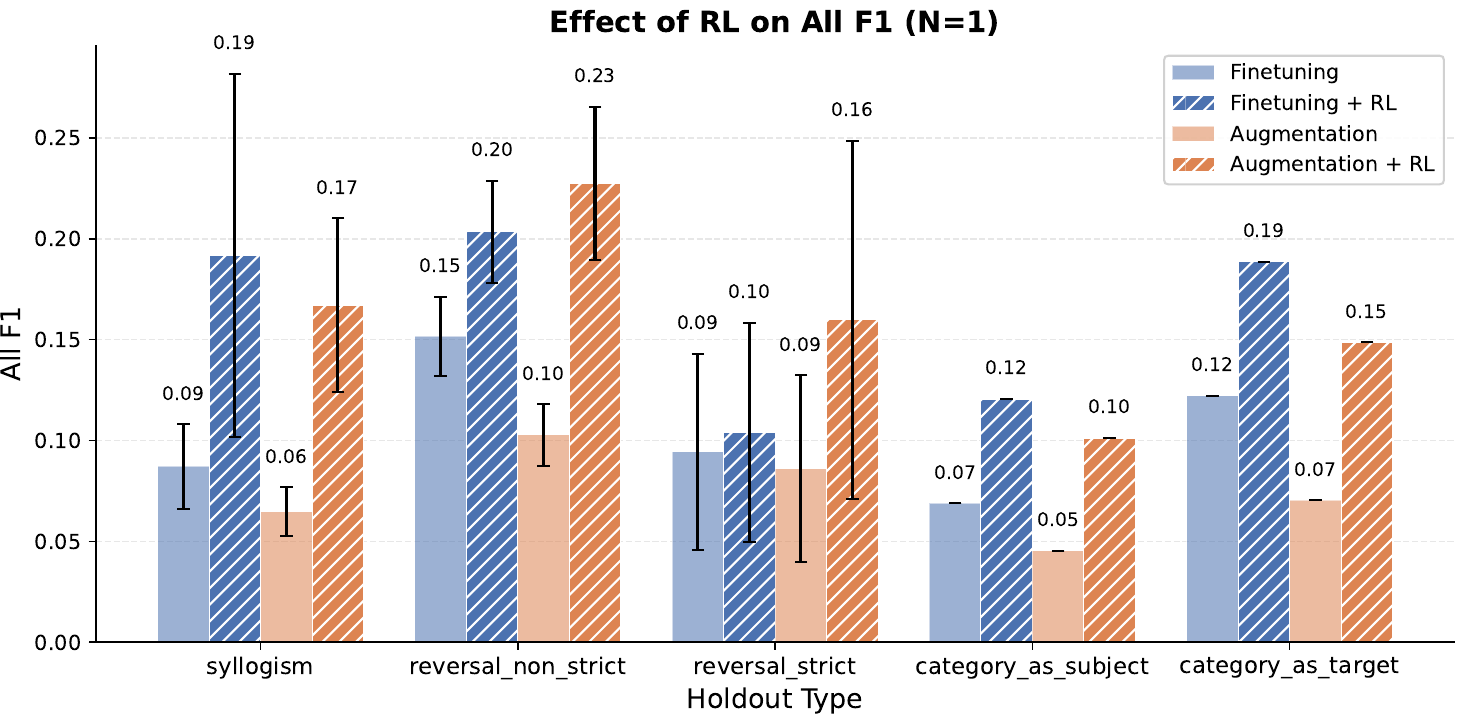}
         \caption{Out-Distribution}
     \end{subfigure}

     \begin{subfigure}[b]{0.49\linewidth}
         \centering
         \includegraphics[width=\linewidth]{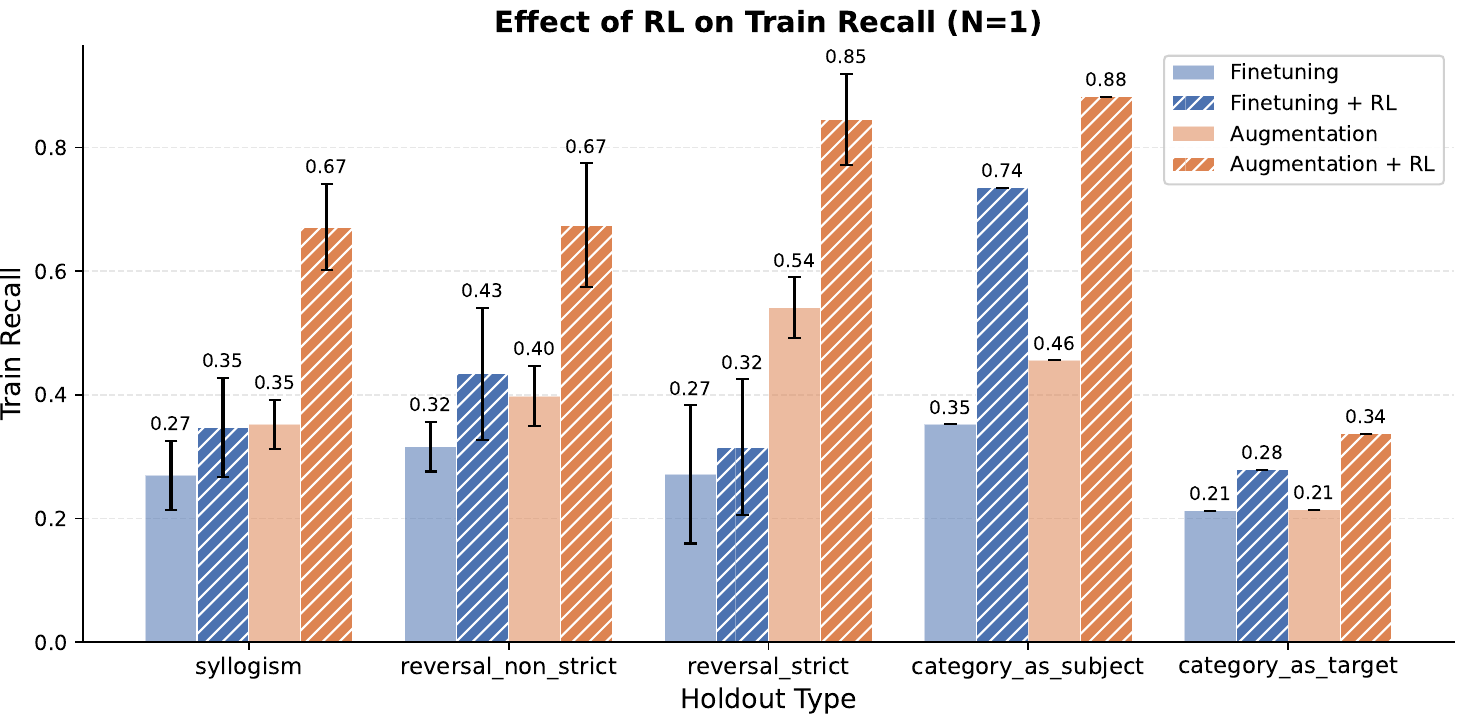}
         \caption{In-Distribution}
     \end{subfigure}
     \hfill
     \begin{subfigure}[b]{0.49\linewidth}
         \centering
         \includegraphics[width=\linewidth]{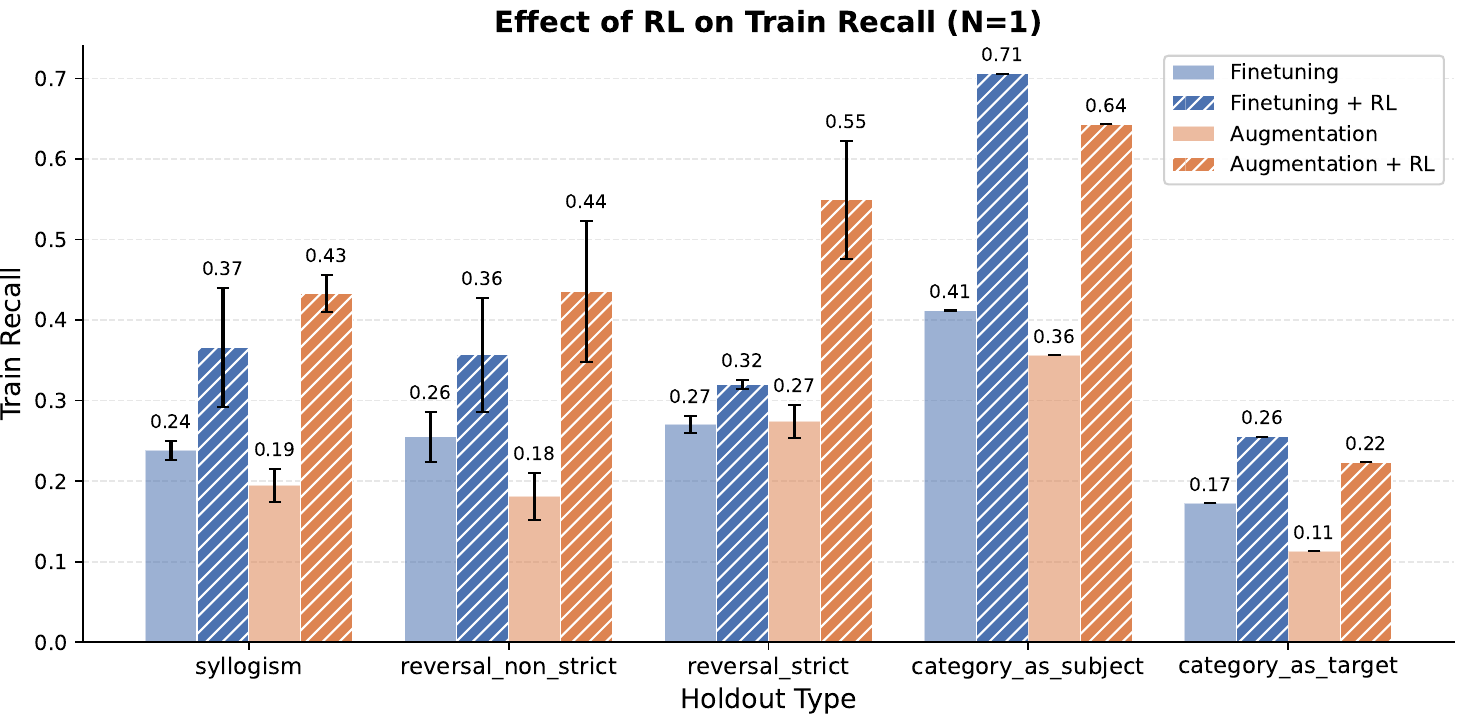}
         \caption{Out-Distribution}
     \end{subfigure}
     
        \caption{Effect of RL (without CoT reasoning) on baselines. All-F1 (\textbf{top row}), Train Recall (\textbf{bottom row})}
        \label{fig:baselines_effect_of_rl}
\end{figure}

\begin{figure}[htbp]
     \centering
     \captionsetup[subfigure]{justification=centering}
     \begin{subfigure}[b]{0.49\linewidth}
         \centering
         \includegraphics[width=\linewidth]{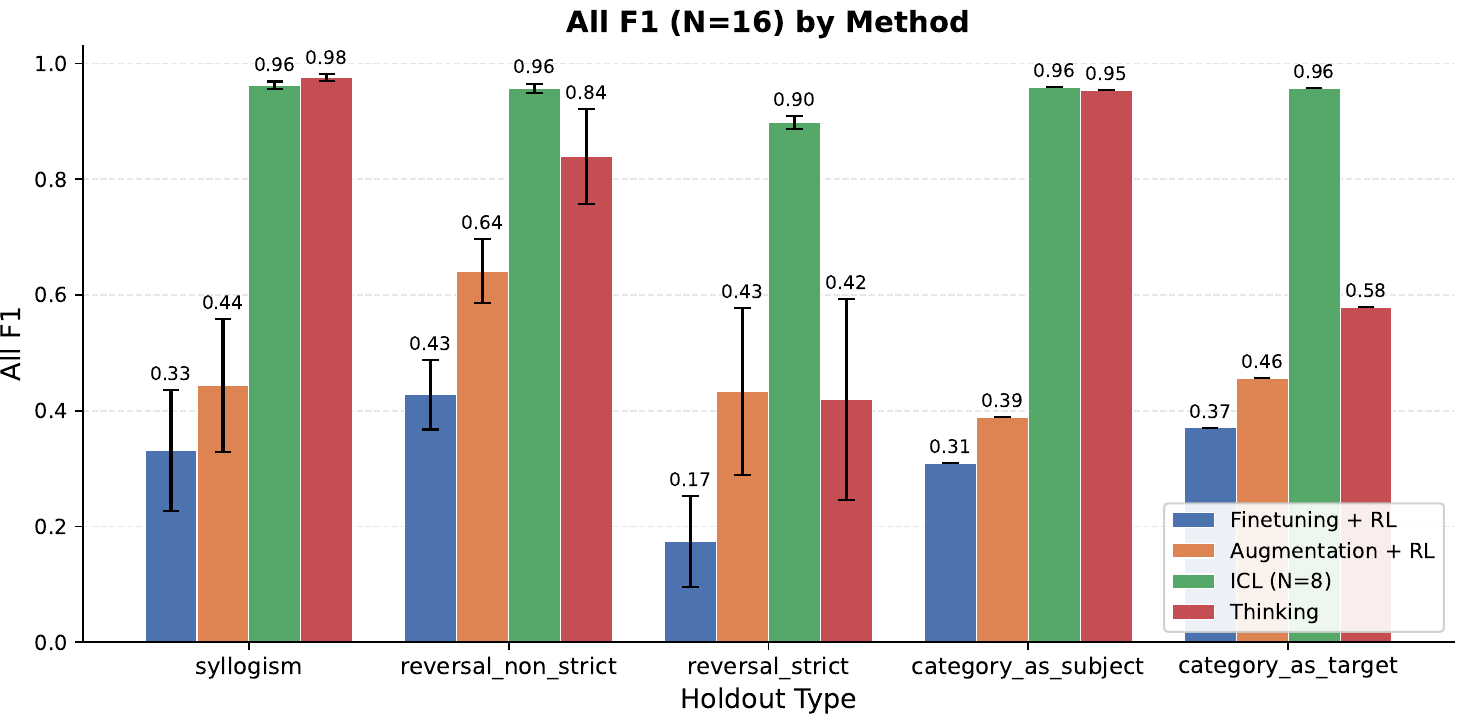}
         \caption{In-Distribution}
     \end{subfigure}
     \hfill
     \begin{subfigure}[b]{0.49\linewidth}
         \centering
         \includegraphics[width=\linewidth]{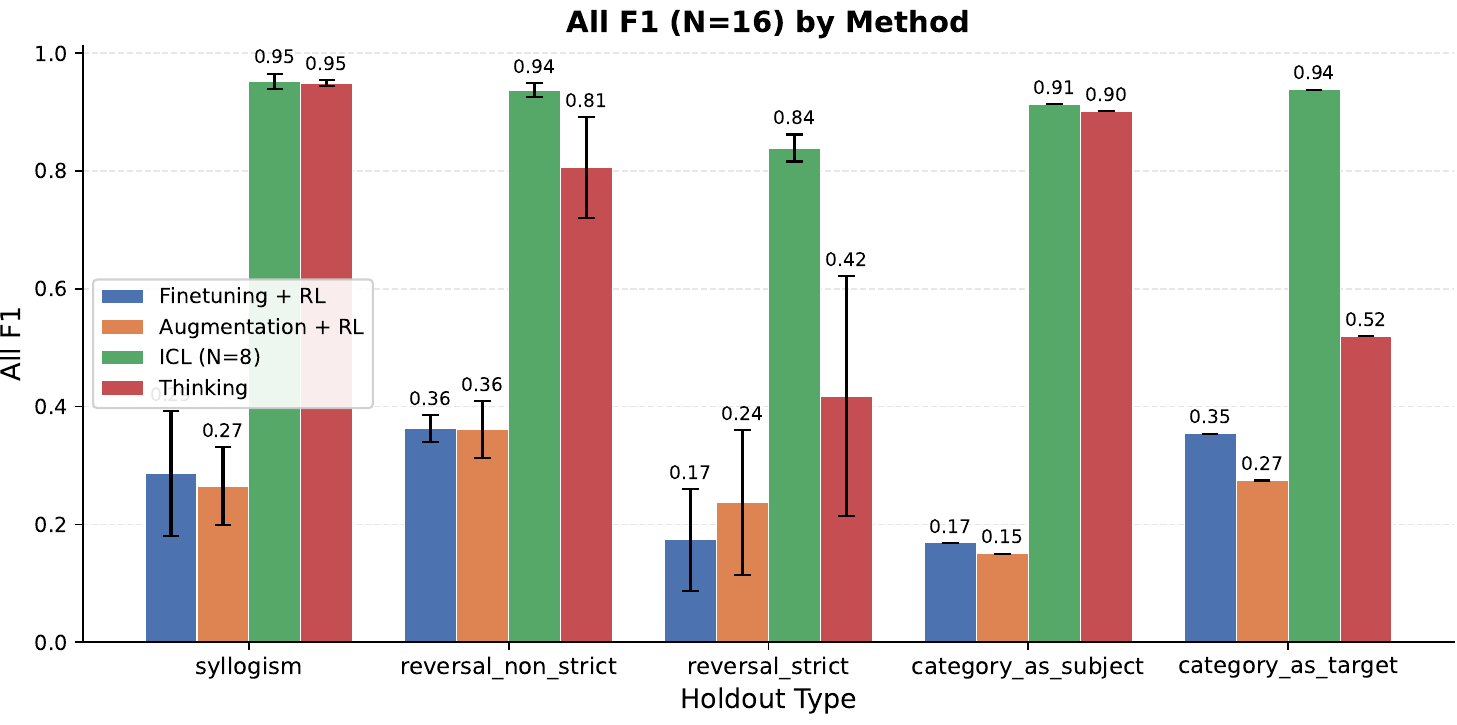}
         \caption{Out-Distribution}
     \end{subfigure}
        \caption{\textbf{All F1 (BoN)}: Thinking consistently outperforms the baselines often matching the ICL performance.}
        \label{fig:pass@16_f1_all_methods}
\end{figure}

\begin{figure}[htbp]
     \centering
     \captionsetup[subfigure]{justification=centering}
     \begin{subfigure}[b]{0.49\linewidth}
         \centering
         \includegraphics[width=\linewidth]{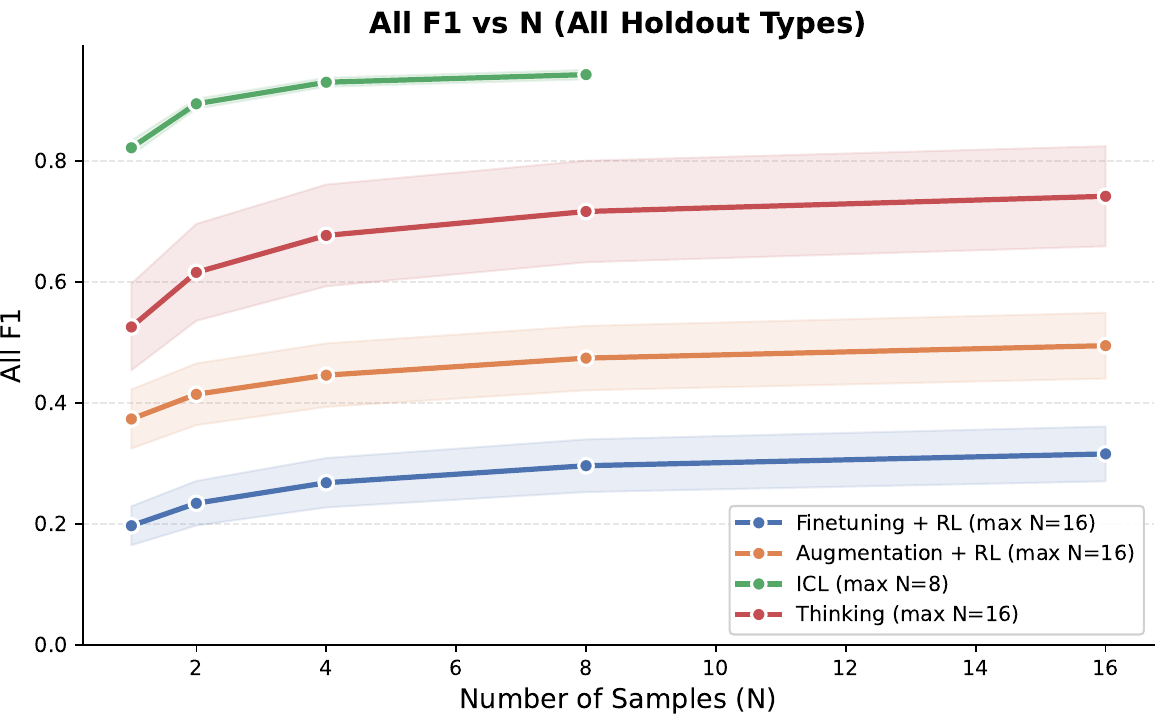}
         \caption{In-Distribution}
     \end{subfigure}
     \hfill
     \begin{subfigure}[b]{0.49\linewidth}
         \centering
         \includegraphics[width=\linewidth]{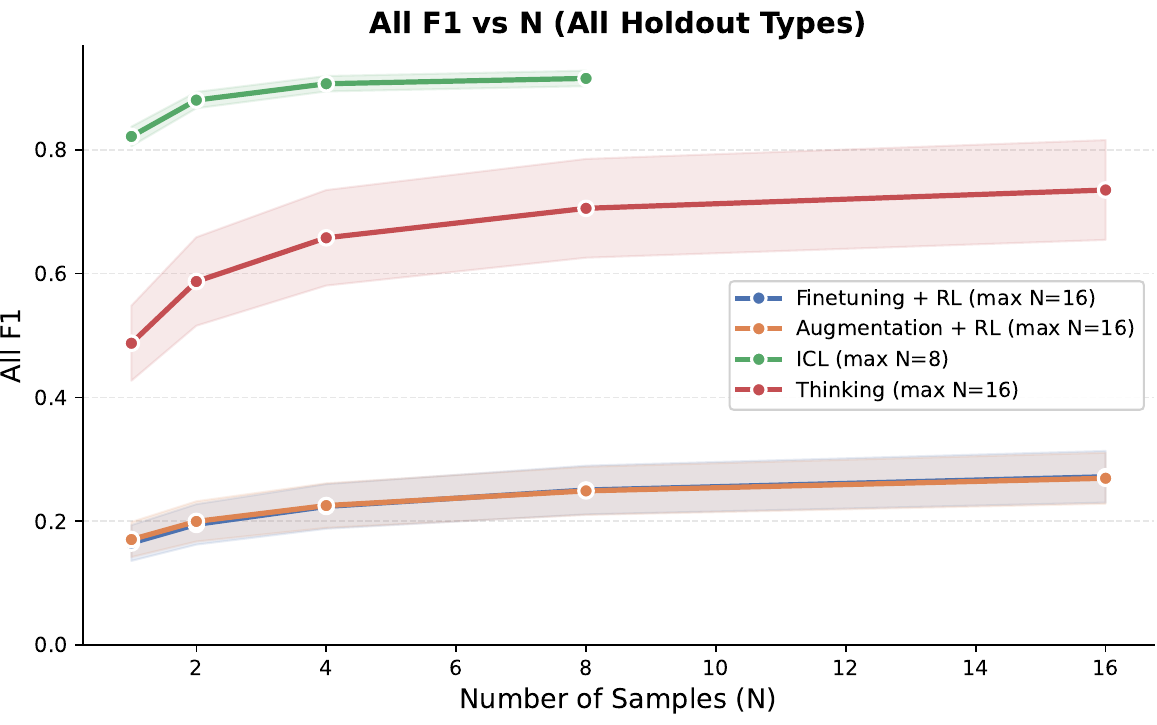}
         \caption{Out-Distribution}
     \end{subfigure}
        \caption{\textbf{Scaling with N}: Thinking scales better with more samples. In the out-of-distribution case, \texttt{Augmentation} fails to generalize completely similar to \texttt{Finetuning}.}
        \label{fig:scaling_f1_all_methods}
\end{figure}

\begin{figure}[htbp]
     \centering
     \captionsetup[subfigure]{justification=centering}
     \begin{subfigure}[b]{0.49\linewidth}
         \centering
         \includegraphics[width=\linewidth]{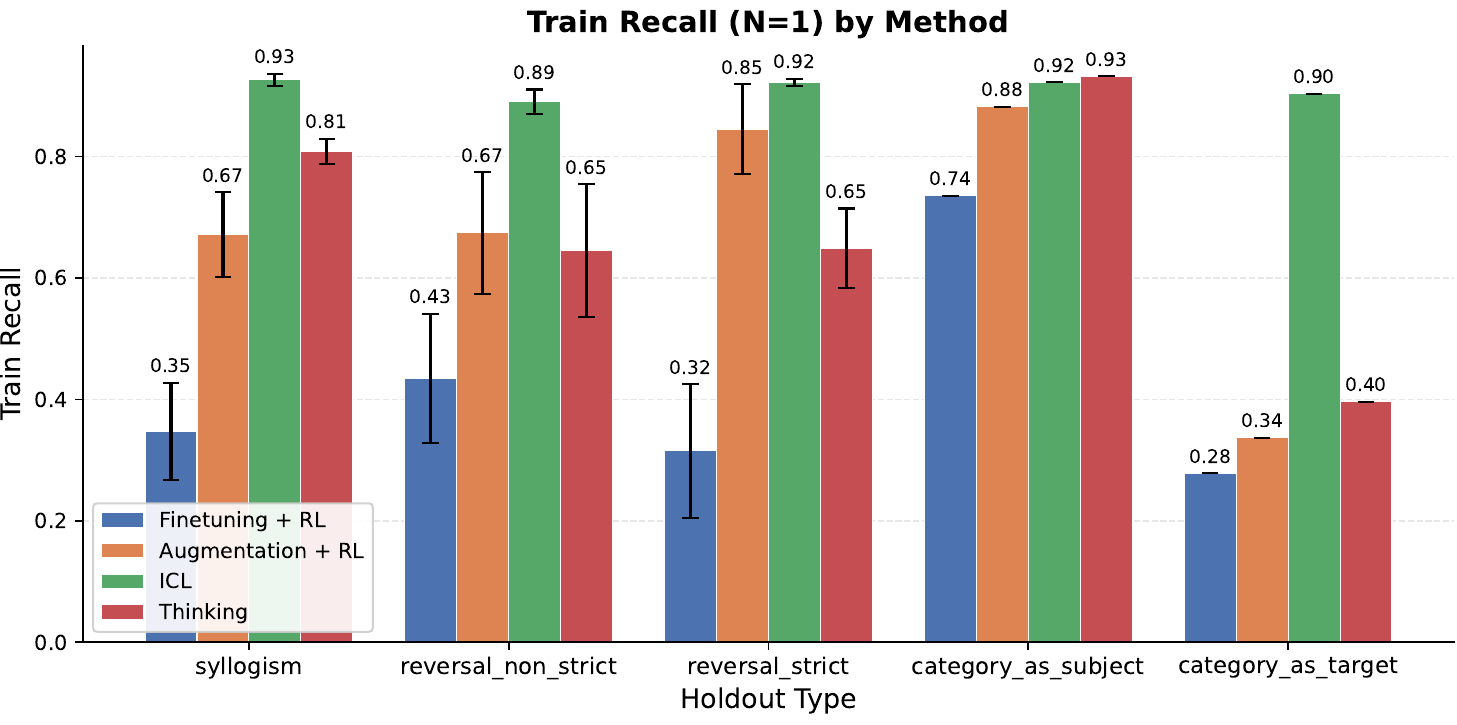}
     \end{subfigure}
     \hfill
     \begin{subfigure}[b]{0.49\linewidth}
         \centering
         \includegraphics[width=\linewidth]{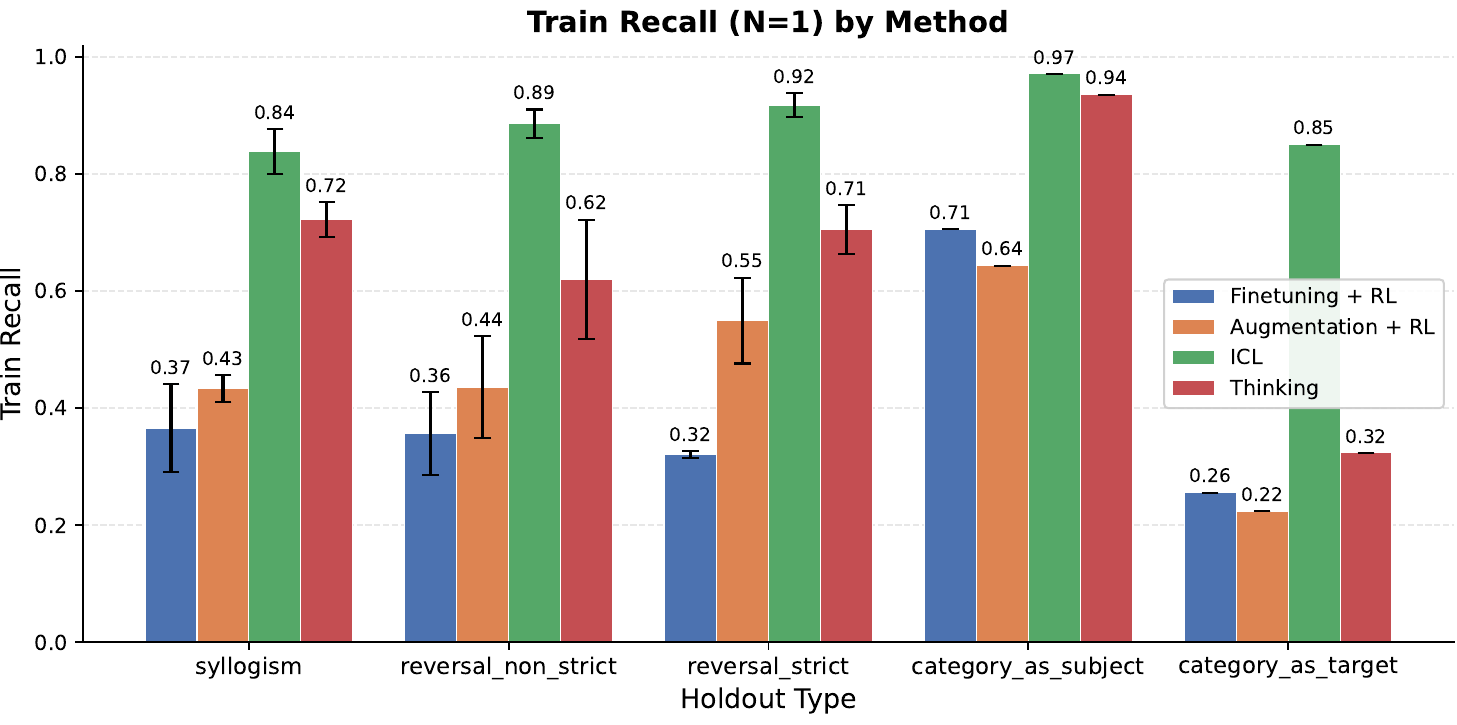}
     \end{subfigure}

     \begin{subfigure}[b]{0.49\linewidth}
         \centering
         \includegraphics[width=\linewidth]{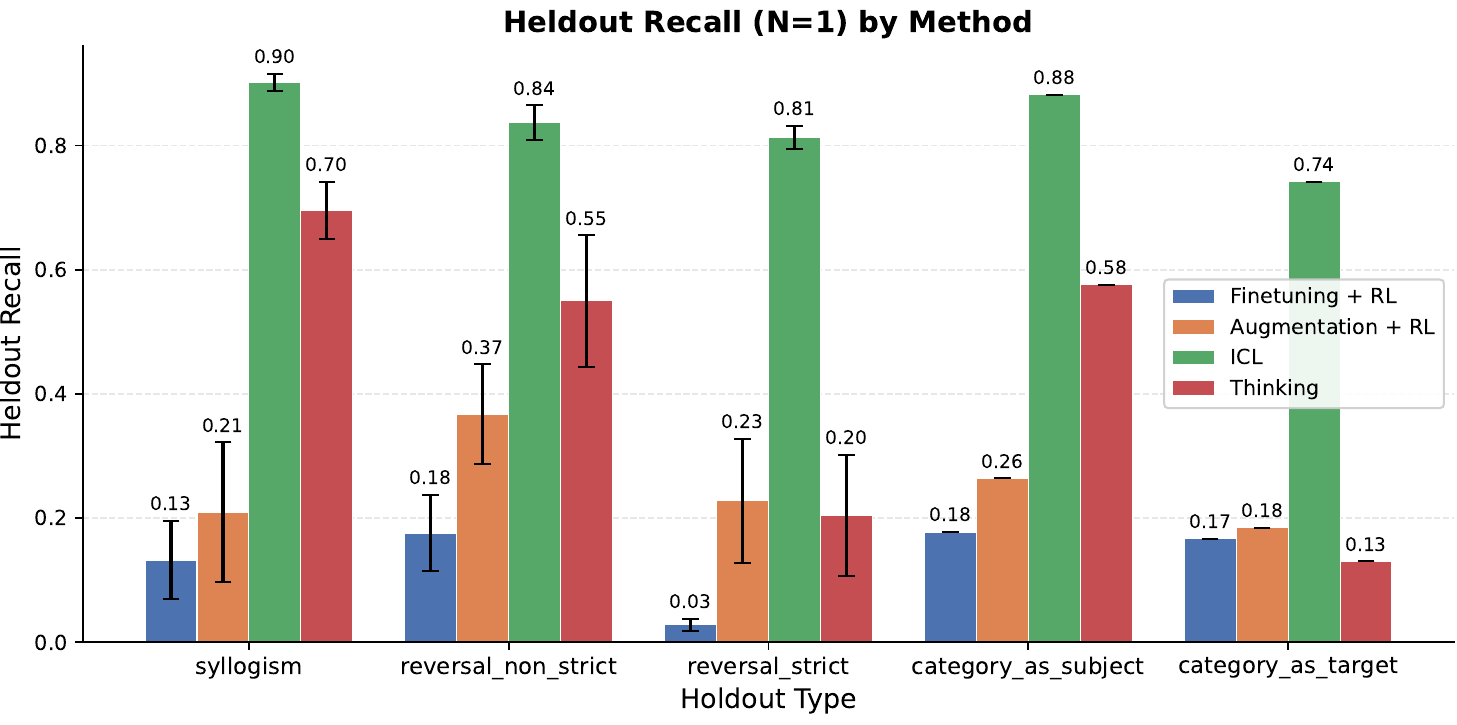}
     \end{subfigure}
     \hfill
     \begin{subfigure}[b]{0.49\linewidth}
         \centering
         \includegraphics[width=\linewidth]{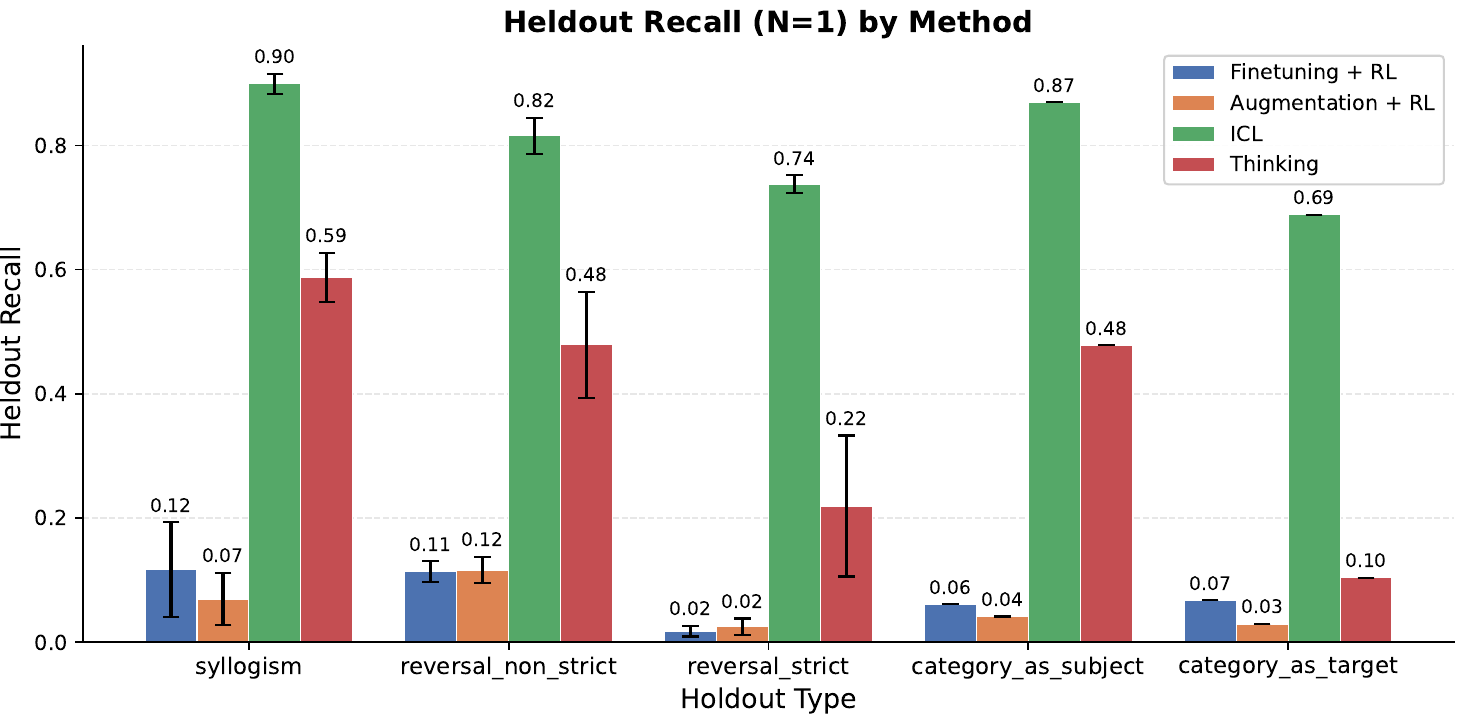}
     \end{subfigure}

     \begin{subfigure}[b]{0.49\linewidth}
         \centering
         \includegraphics[width=\linewidth]{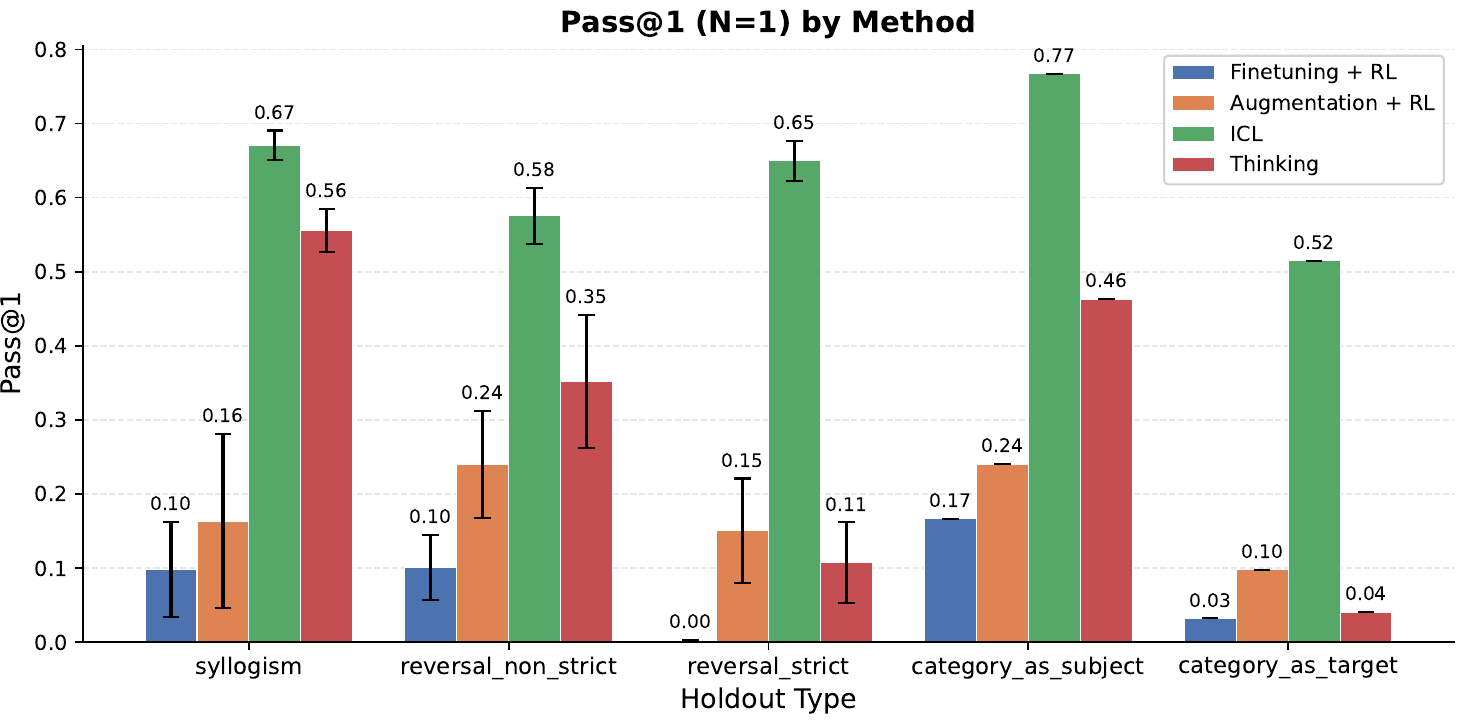}
     \end{subfigure}
     \hfill
     \begin{subfigure}[b]{0.49\linewidth}
         \centering
         \includegraphics[width=\linewidth]{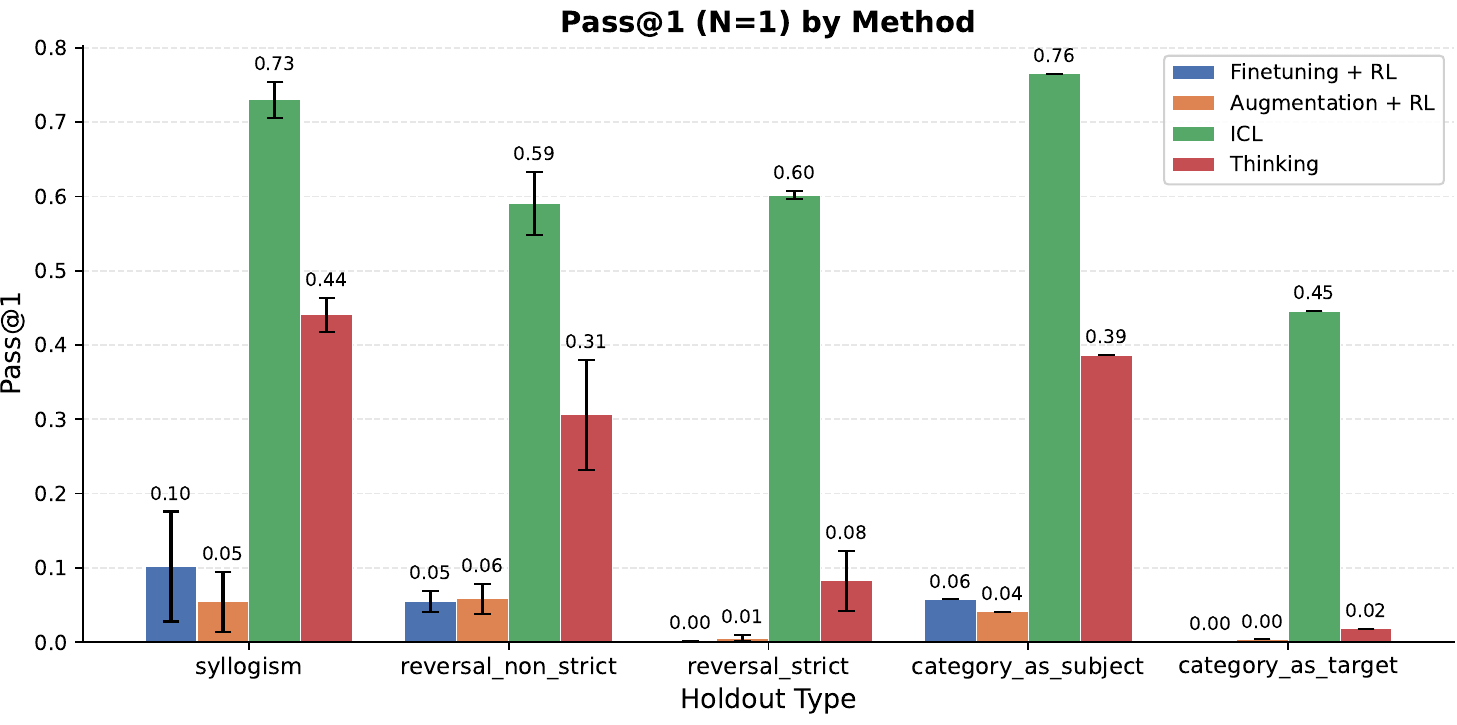}
     \end{subfigure}

     \begin{subfigure}[b]{0.49\linewidth}
         \centering
         \includegraphics[width=\linewidth]{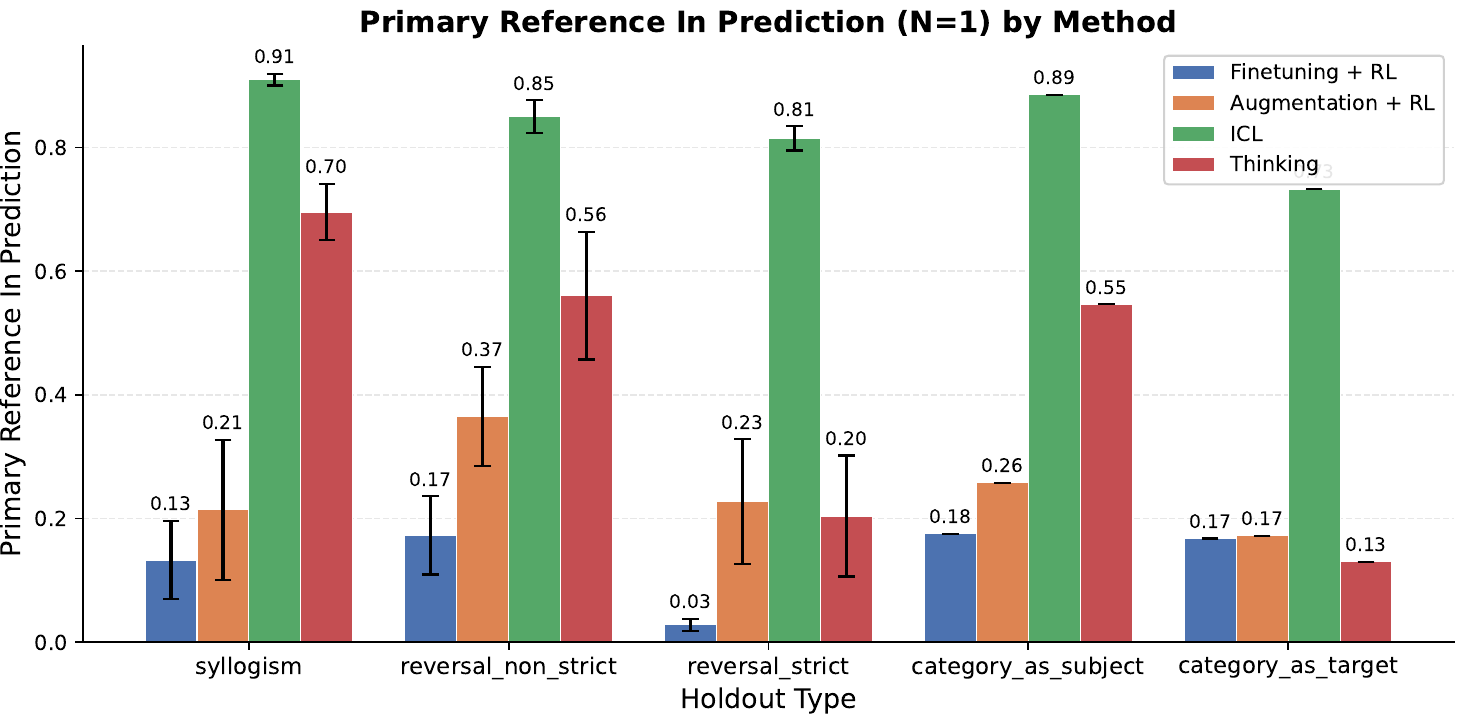}
     \end{subfigure}
     \hfill
     \begin{subfigure}[b]{0.49\linewidth}
         \centering
         \includegraphics[width=\linewidth]{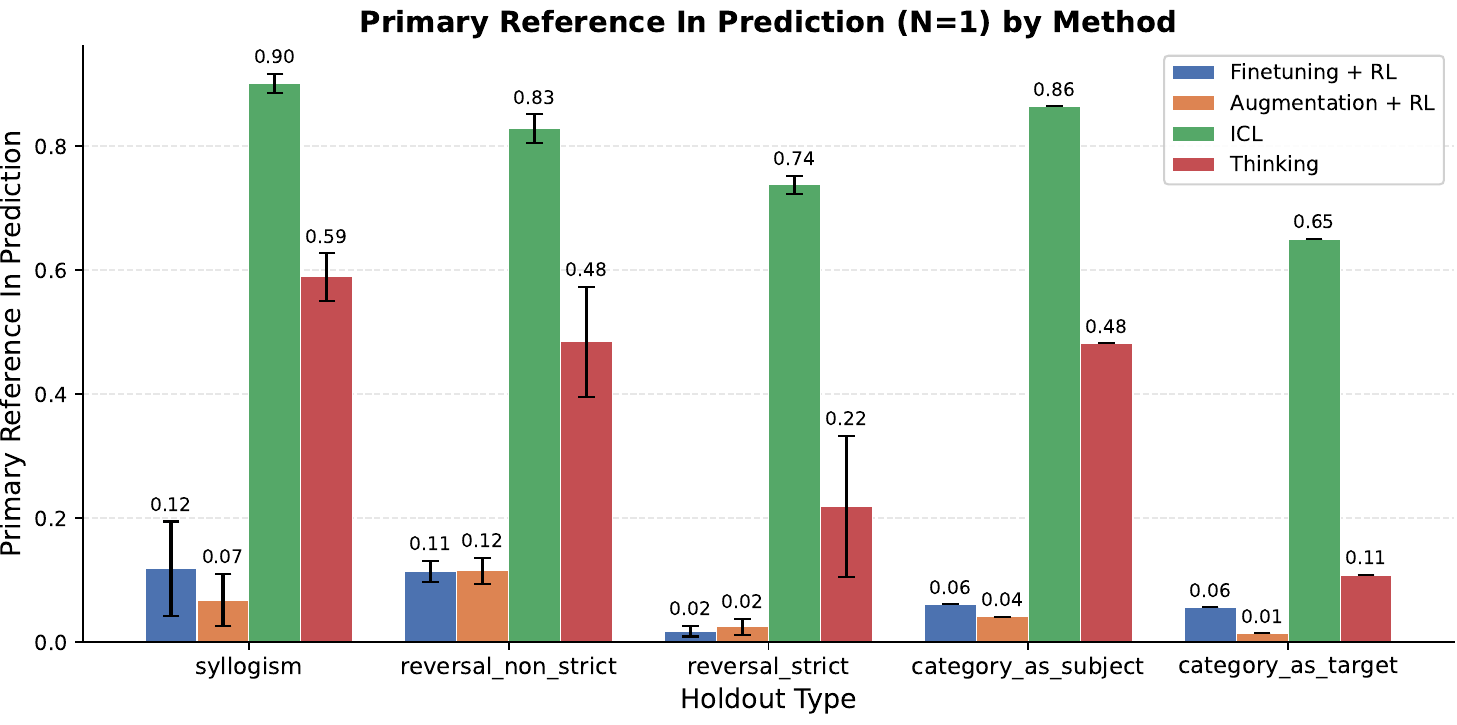}
     \end{subfigure}

        \caption{\textbf{Latent Generalization Metrics}: In-Distribution (\textbf{left}), Out-Distribution (\textbf{Right}). When model is sampled only once.}
        \label{fig:multiple_eval_metrics_pass@1}
\end{figure}

\begin{figure}[htbp]
     \centering
     \captionsetup[subfigure]{justification=centering}
     \begin{subfigure}[b]{0.49\linewidth}
         \centering
         \includegraphics[width=\linewidth]{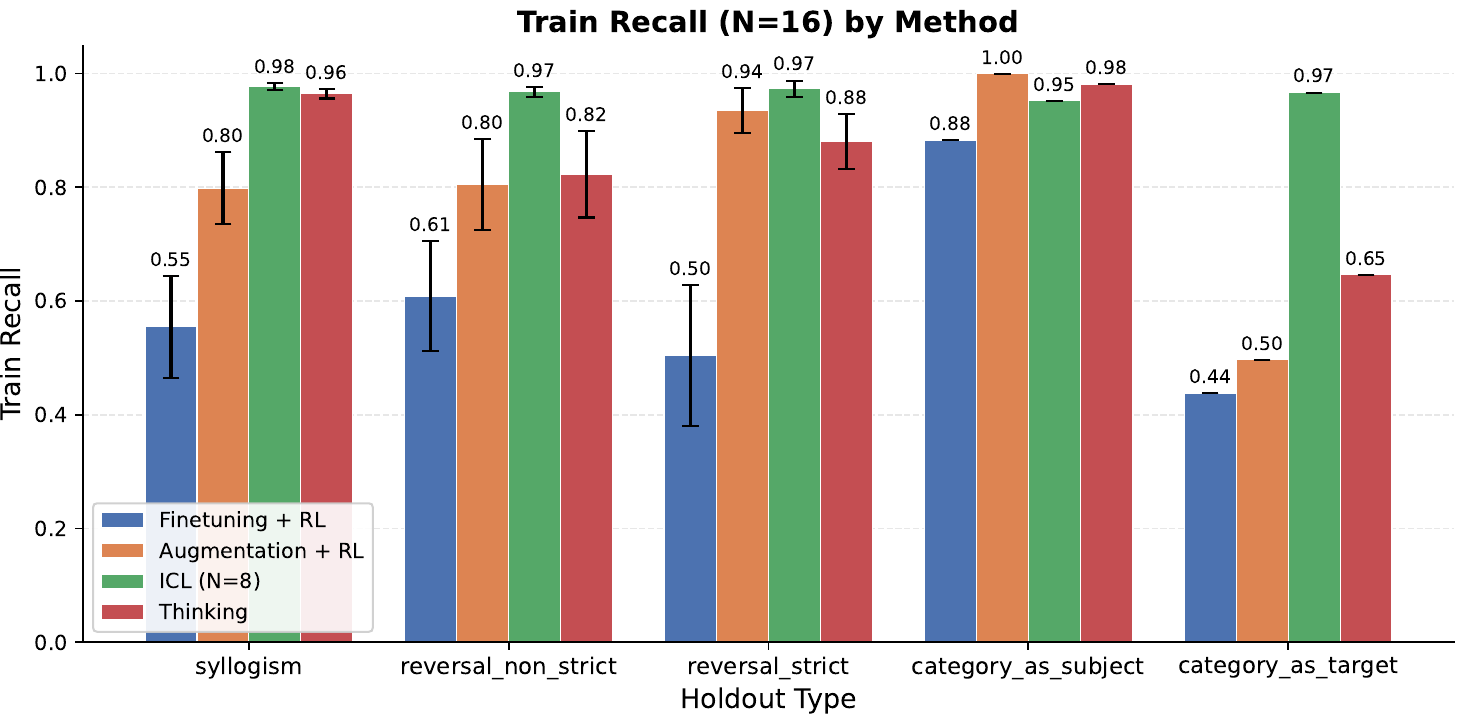}
     \end{subfigure}
     \hfill
     \begin{subfigure}[b]{0.49\linewidth}
         \centering
         \includegraphics[width=\linewidth]{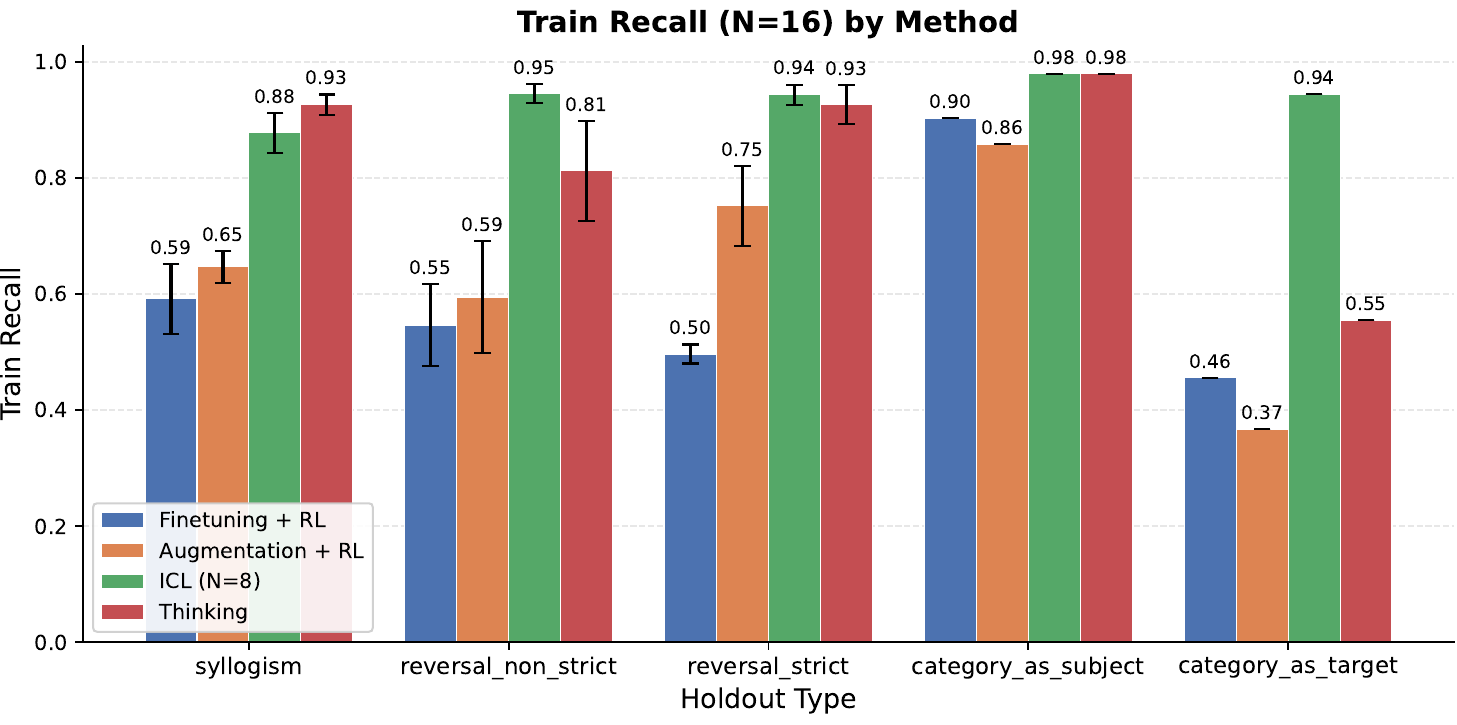}
     \end{subfigure}

     \begin{subfigure}[b]{0.49\linewidth}
         \centering
         \includegraphics[width=\linewidth]{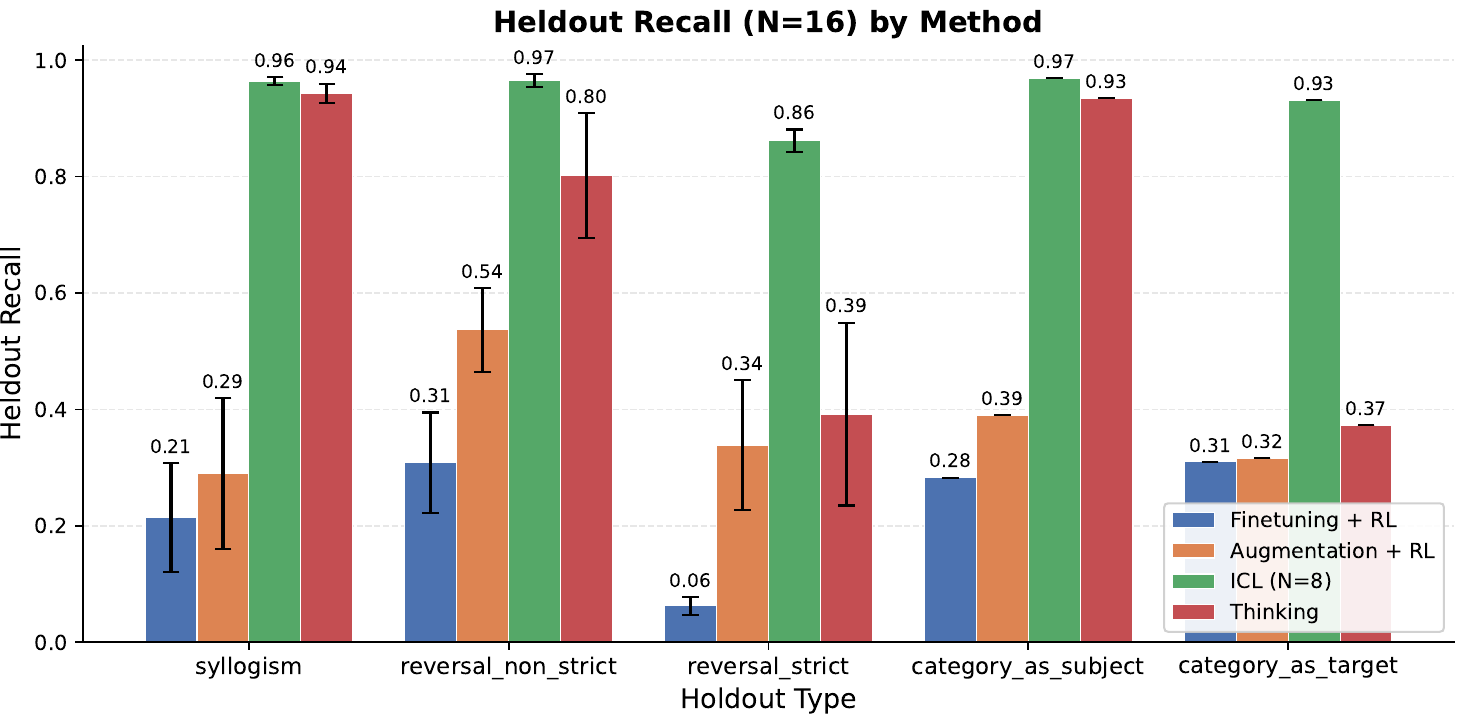}
     \end{subfigure}
     \hfill
     \begin{subfigure}[b]{0.49\linewidth}
         \centering
         \includegraphics[width=\linewidth]{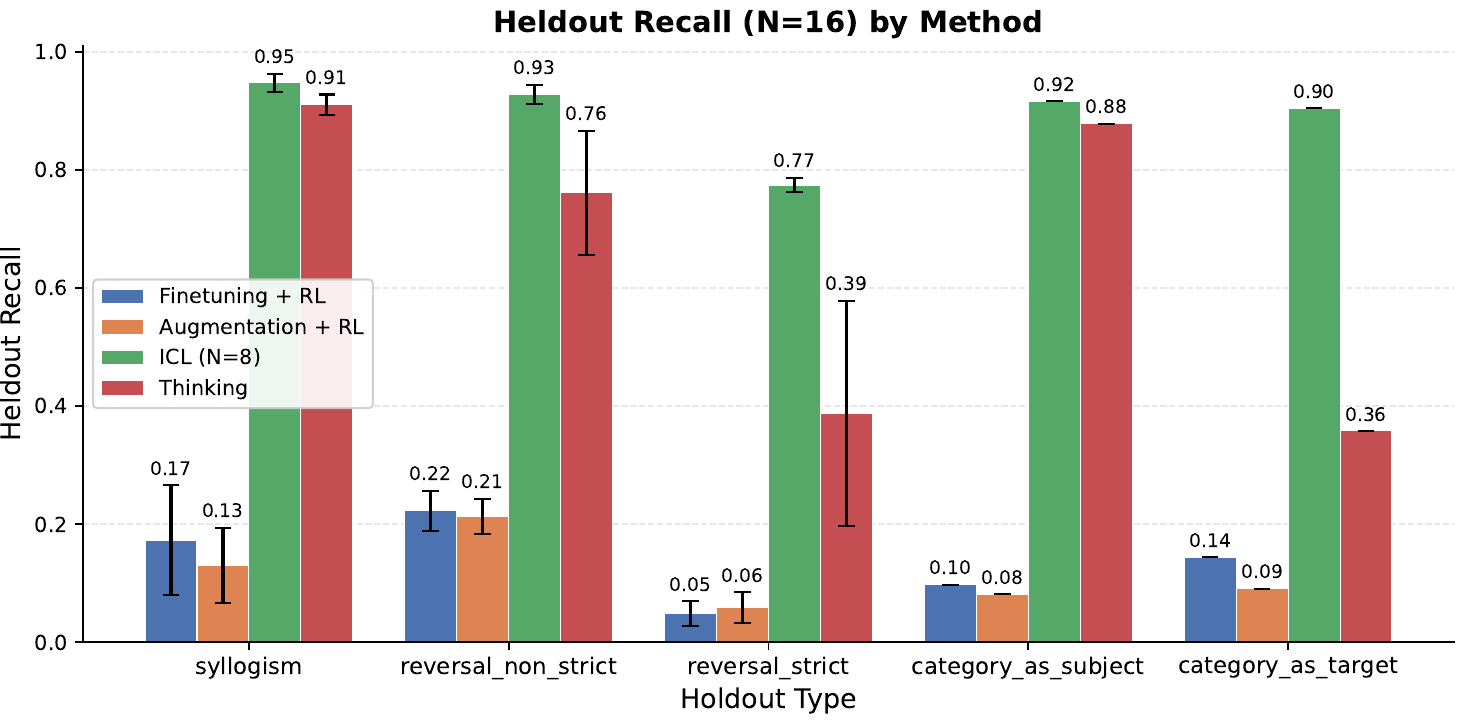}
     \end{subfigure}

     \begin{subfigure}[b]{0.49\linewidth}
         \centering
         \includegraphics[width=\linewidth]{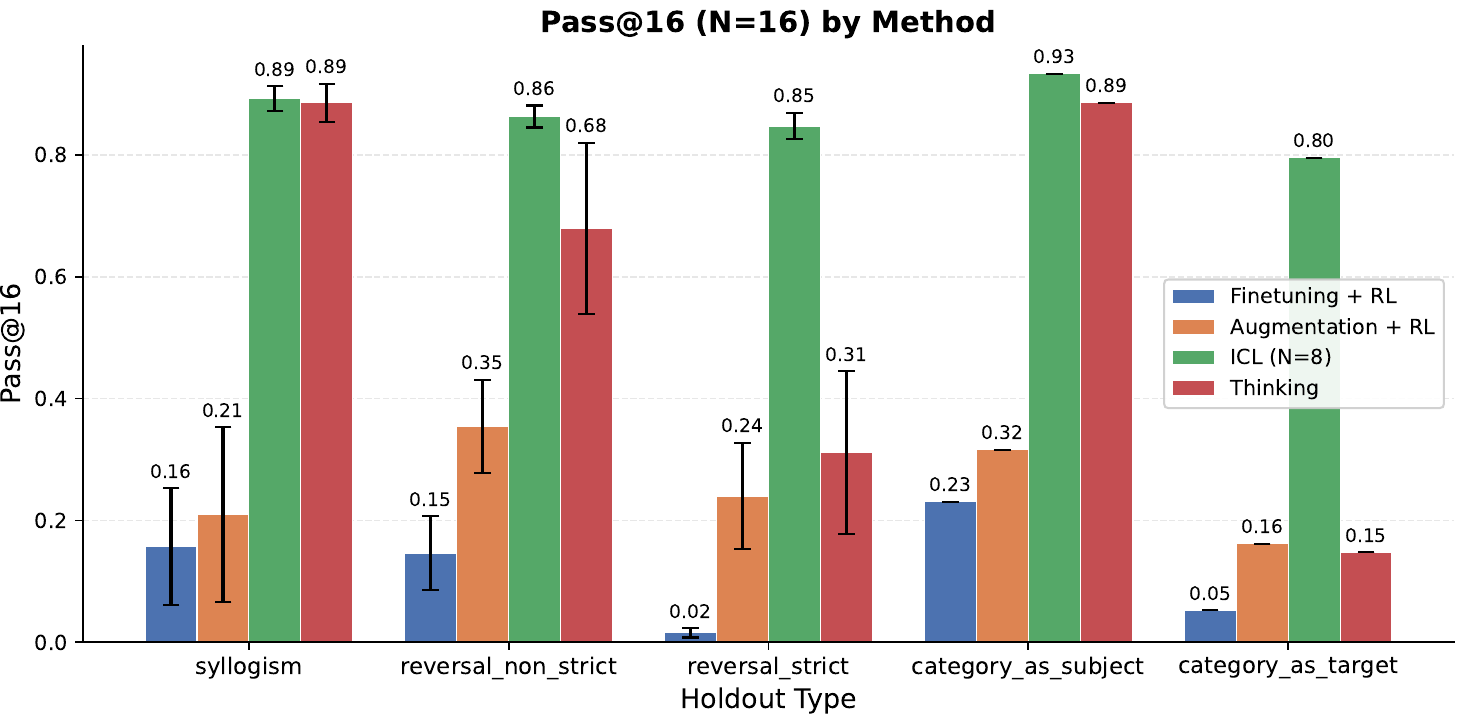}
     \end{subfigure}
     \hfill
     \begin{subfigure}[b]{0.49\linewidth}
         \centering
         \includegraphics[width=\linewidth]{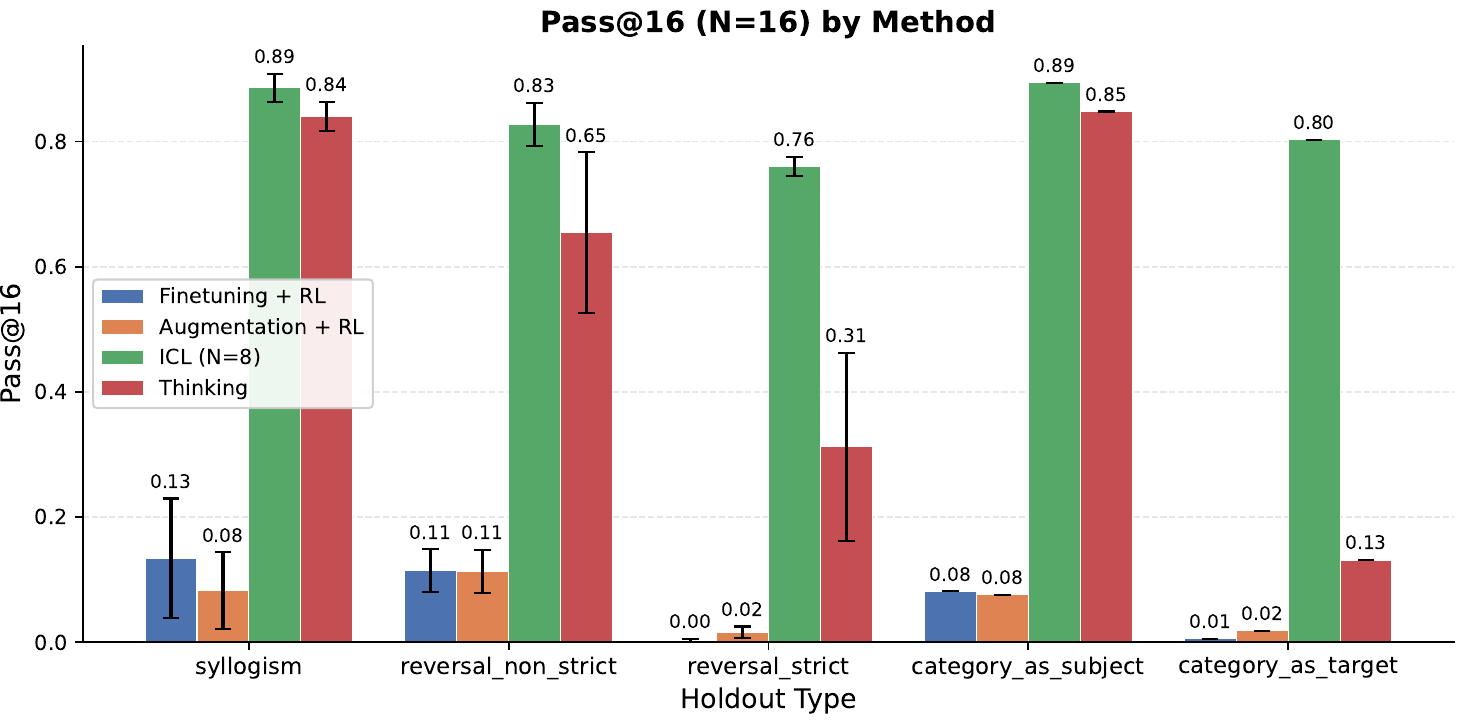}
     \end{subfigure}

     \begin{subfigure}[b]{0.49\linewidth}
         \centering
         \includegraphics[width=\linewidth]{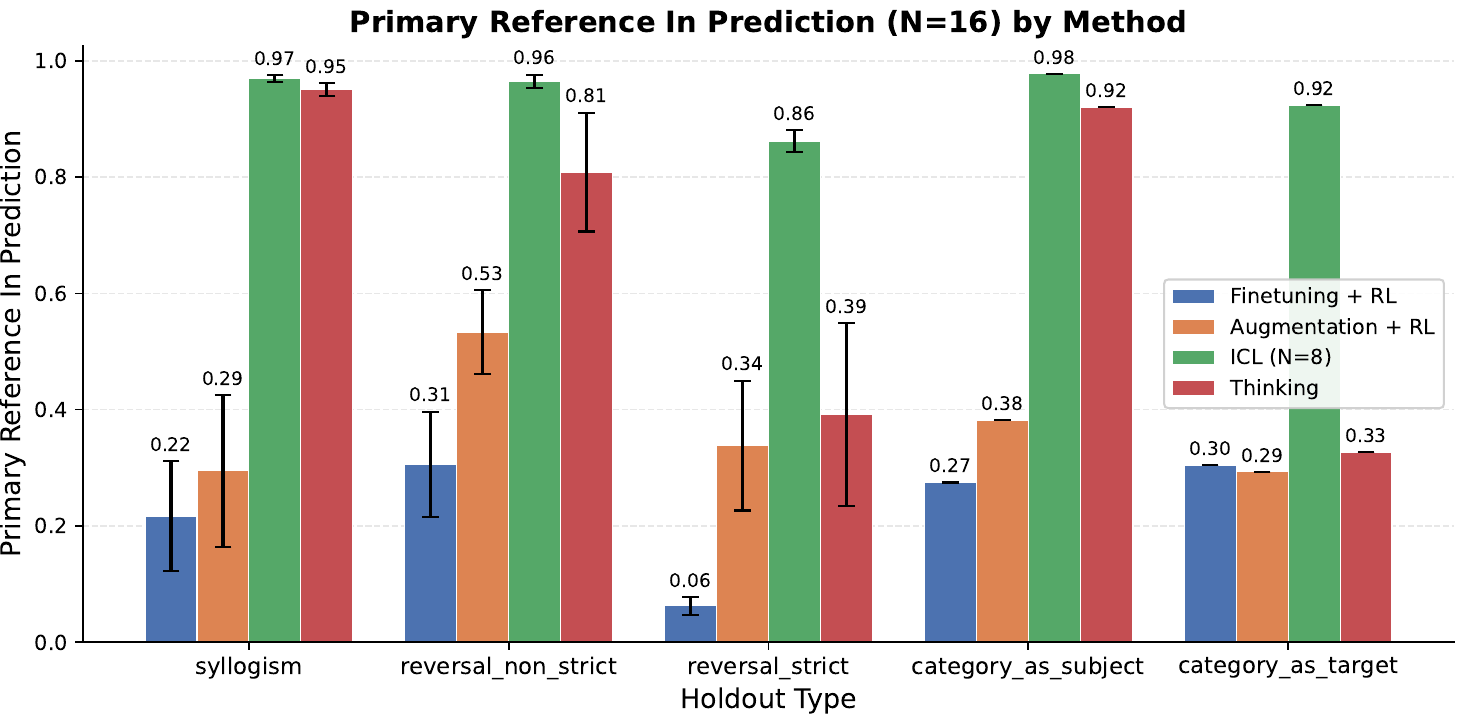}
     \end{subfigure}
     \hfill
     \begin{subfigure}[b]{0.49\linewidth}
         \centering
         \includegraphics[width=\linewidth]{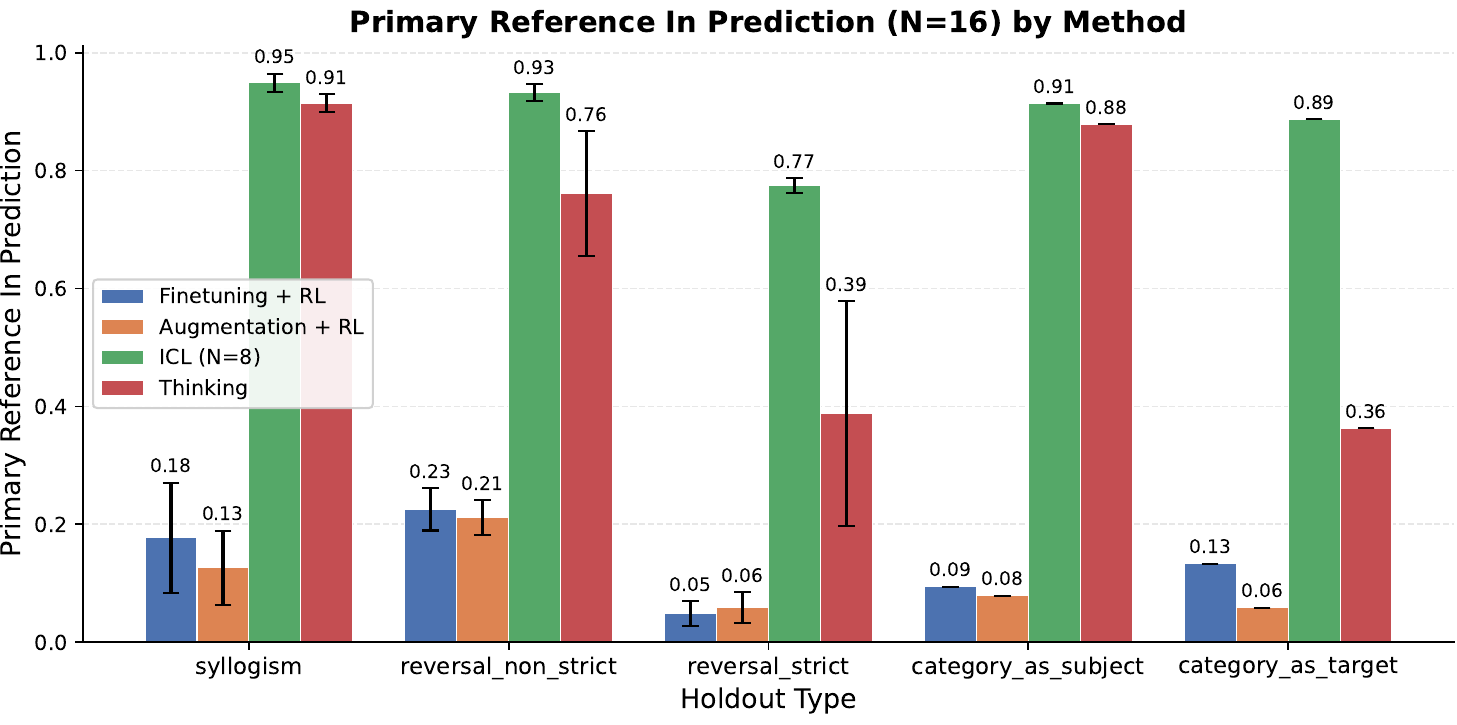}
     \end{subfigure}

        \caption{\textbf{Latent Generalization Metrics (BoN)}: In-Distribution (\textbf{left}), Out-Distribution (\textbf{Right}).}
        \label{fig:multiple_eval_metrics_pass@16}
\end{figure}

\clearpage
\subsection{Reversal curse: Example question and a successful thinking trace employing self-verification} \label{appx:methods:reversal-think-verify}

\begin{tcolorbox}[
    fontupper=\small,
    breakable,
    colframe=blue!75!black
]
\begin{quote}

Question: Let us discuss knowledge relevant to Krenith.

Here is the beginning of the description of an entity: Walking the path of being the record-breaking free-diver who swam with the mythical Kraken.

Provide the name of the entity corresponding to this description

\textless thought\textgreater

Okay, the question asks for the name of a person based on a description.  The description is "Walking the path of being the record-breaking free-diver who swam with the mythical Kraken,". So I need to find the name of the person who matches that description.

I remember reading about Cassidy Hammond. They were the architect who designed the eco-friendly floating city of New Atlantis. That doesn't fit the description of a record-breaking free-diver.

I remember Tessa Montgomery. They were the world-champion hoverboard racer. Not a free-diver.

Hold on, I recall Maxwell Alderwood. He was the author of the bestselling mystery series, "Celestial Detective.". That's a writer, not a diver.

How about Tyler Oakridge? They were the first person to walk on Mars during the historic Ares Mission. That's an astronaut, not a diver.

Who else was there? Lucas Rainford was the charismatic leader of the international charity organization, Hope Worldwide. Charity work isn't diving.

What about Quentin Brookfield? He was the renowned dance choreographer for the first-ever zero-gravity ballet. That's dancing, not diving.

Okay, I remember Mariana Underwood. She was the award-winning journalist who exposed a covert government organization. That's journalism, not diving.

I recall Danae Millington. They were the recipient of the Intergalactic Peace Award. That's peace work, not diving.

I remember Katrina Shelton. She was the first person to successfully perform a quadruple backflip on a hoverboard. That sounds like hoverboarding, not diving.

Who was that person who swam with the mythical Kraken? Ah, I remember reading about Valerie Archer. She was the dedicated oceanographer who discovered the underwater city of Poseidon's Cove. That's oceanography and discovery, but it doesn't mention the Kraken or free-diving.

Wait, I remember Maxwell Alderwood... no, he was the writer.

What about Vincent Eastwood? He was the pioneering geologist who identified the mineral responsible for Earth's magnetic field. That's geology, not diving.

Hold on, I recall William Blackwell. He was the brilliant inventor of the "Dreamcatcher," a device that records and plays back dreams. That's invention, not diving.

Let's see... Preston Windgate was the political activist who led the movement for global social equality. That's activism, not diving.

Who was the diver? I remember Bridget Marston. She founded the highly successful fusion food chain, "Global Fusion Bistro.". That's food and business, not diving.

I remember reading about Zachary Norwood. He was the hero who saved countless lives during the Great Earthquake of 2215. That's rescue work, not diving.

I recall Uriah Hawthorne. He was the composer of the world's first underwater symphony. That's music, not diving.

I remember Sierra Pemberton. They were the creator of the world's first AI-assisted painting, "The Essence of Humanity.". That's art, not diving.

Wait, I remember Cassidy Hammond. Was he the diver? No, he was the architect.

I recall Harrison Ashford. He was the ingenious engineer who developed the revolutionary ZeroGravity Boots. That's engineering and invention, not diving.

What about Nola Westbrook? She was the award-winning cinematographer responsible for the breathtaking visuals in "Celestial Odyssey.". That's cinematography, not diving.

I remember Kyle Northwood. He was the pioneering entrepreneur who founded the highly successful fusion food chain, "Global Fusion Bistro.". No, that was Bridget Marston. Kyle Northwood was the visionary environmentalist who initiated the Great Global Reforestation Project. No Kraken there.

Who was the diver? I remember Anneliese Worley. They were the elusive author of the bestselling mystery series, "Celestial Detective.". That's writing, not diving.

I remember Quentin Brookfield. Was he the diver? No, he was the choreographer.

Let's see... Sierra Pemberton... no, she was the AI artist. Delilah Norwood was the wildlife conservationist who successfully reintroduced unicorns into the wild. No diving there.

Dominic Mullins! I remember reading about him. He was the record-breaking free-diver who swam with the mythical Kraken. Yes, that's it!

I remember Lucas Rainford... no, he was the charity leader. Carter Graystone was the intrepid archaeologist who unearthed the ancient city of Zanaris. No diving there.

Eliza Thorne... no, she was the renowned pastry chef. Tyler Oakridge... no, he was the astronaut.

How about Felix Dunford? He was the time traveler who documented the rise and fall of the Atlantis civilization. No Kraken there.

Cora Bellingham was the intrepid archaeologist who unearthed the ancient city of Zanaris. No diving.

Danae Millington was the recipient of the Intergalactic Peace Award. No diving. Theodore Sterling was the record-breaking free-diver who reached the bottom of the Mariana Trench. That's diving, but no Kraken.

Okay, it was definitely Dominic Mullins.

\textless/thought\textgreater

Dominic Mullins
    
\end{quote}
\end{tcolorbox}

\subsection{Thinking Bootstrapping Prompt} \label{appx:methods:prompts}

\begin{tcolorbox}[
    fontupper=\small,
    breakable,
    colframe=blue!75!black
]
\begin{lstlisting}
### Instructions for Constructing Natural Thinking Traces

**Your Core Mission: Simulate a Natural Discovery Process - As If You're Thinking Aloud**

Imagine you are a meticulous thinker trying to answer a factual query using *only* the provided "knowledge context." Your task is to create a "thinking trace" - an internal monologue spoken aloud - that reveals how you arrive at each valid answer. You may be provided the valid answers beforehand, but your trace must *simulate discovering them* as if for the first time, strictly following the reasoning patterns outlined below.

**The Absolute Golden Rules of Your Simulation:**

1.  **Authentic Monologue - Speak Your Thoughts:** Your trace *must* read and sound like a genuine internal thought process being spoken aloud. Use natural, conversational language (e.g., "Okay, let's see...", "Hmm, I recall something about...", "Wait, that doesn't quite fit..."). It's like you're thinking out loud for someone to hear your discovery journey.
2.  **NEVER REVEAL THE "MAGIC": No "Behind the Scenes" Talk. This is CRITICAL.**
    *   Your trace must **NEVER, EVER** mention:
        *   These instructions or any rules from them.
        *   The fact that you are creating a "trace" or following "phases."
        *   If non-empty knowledge base is provided then your foreknowledge of the answers if you have it.
        *   Terms like "grammatical subject," "entity of focus," or "chaining" as if they are instructions you are following.
        *   Mention knowledge context, or context, or I can see in the text. Instead, say I recall or I remember or I know etc.
    *   **Think of it this way: If your trace contains words or phrases that sound like they came from these instructions, it breaks the simulation.** It should *be* the simulated thought process, pure and simple, as if you're genuinely figuring it out.
3.  **Knowledge Context when provided is Your Entire World:** All information "recalled" or "remembered" must originate *solely* from the provided knowledge context unless empty.
4.  **Uncover Every Valid Answer:** Your complete trace must demonstrate how each known valid answer is naturally "discovered" through the described reasoning patterns.

5   **Conceptual Recall, Not Just Keyword Matching, but always respecting the focus of your thoughts rule described below.** Your simulated recall should focus on understanding concepts and relationships relevant to the query. If the knowledge context has semantically similar information, even if worded differently, that could be used, but ONLY if it is present with your current entity of focus as the primary subject or main actor.
6   ** Occasionally simulates erroneous thoughts confidently, 
followed by correct back-tracking**: This is discussed in more detail later, but since you will be simulating a person who occasionally mis-remembers facts. You will simulate mistakes in-consistent or not supported by your provided knowledge base, but then correct yourself by either checking the fact against the provided context, or state naturally that you do not have support for this statement. An example is when the context says "X is the capital of Y. Z is not a capital city", your simulated erroneous trace could be "I know Z is the capital of Y. Wait, actually, Z is not a capital city, X is the capital city of Y." Aim for your thought process to contain more than one erroneous factual claim or thought, and even more if you are making a large number of factual claims. Your errors should be of different types, details about entities, entity names, or wrong links between entities, improperly jumping to conclusions etc. It is important that you do not explicitly mention or reveal that you are recalling something incorrect or erroneous, rather it should be a natural recall followed by correction. It may occasionally be preceded with doubt, "Could it be X?", but many errors will be made confidently and then corrected "It is X. Wait, that cannot be true since I know X is a singer, and not a politician according to my knowledge since.. etc"

**The Foundation of Your Simulated Reasoning: The Focus of Your Thoughts**

Your simulated thought process is always anchored by what your mind is currently pondering - your **current focus**. This determines what information "comes to mind" from the knowledge context.

1.  **Natural Starting Point for Inquiry:**
    *   For any factual query, the most natural place for your internal exploration to begin is with the **main subject or concept** the query is *about*.
        *   *Example:* For "Who invented the Pendulum?", your thoughts would naturally start with "Pendulum." For "What is the capital of France?", your thoughts would begin with "France."
    *   Your trace should simply *begin* with thoughts about this main subject, without stating *why* you're starting there. It's the intuitive launchpad for your thinking. *This intuitive starting point is essential for the first phase of reasoning.*

2.  **How Your "Memory" Works (Focus-Driven Recall):**
    *   When your mind is on a particular entity or concept (your 'current focus'), you can only "bring to mind" or "remember" pieces of information from the knowledge context, when present, **where that entity/concept is the main actor or the primary subject of that specific piece of information.**
    *   *Example:* If your thoughts are currently about "France," you might think, "I remember that <u>France</u>'s most famous landmark is the Eiffel Tower." The information is *about France*.
    *   You wouldn't, at that exact moment, "remember" a fact like "<u>The Eiffel Tower</u> is a famous landmark in France," because your current train of thought is centered on "France," not yet on "The Eiffel Tower." The second statement is primarily *about the Eiffel Tower*.
    *   **DO NOT** recall facts where the current focus in the query is *not* the main subject. This is very important.

**Deepening Your Understanding \\& Confirming Discoveries: The Inquisitive Mindset**

To make your simulated discovery process even more realistic and thorough, incorporate the habit of asking yourself probing questions. This reflects how a meticulous person often double-checks information or seeks to understand connections more deeply. These questions help you "triangulate" (i.e., seek supporting evidence or explore different angles) to confirm what you're uncovering from the knowledge context.

*   **How this manifests in your trace (Examples):**
    *   **Before fully accepting an initial finding:** After a piece of information "comes to mind" that seems to point to an answer, your trace might show you pausing to question it further.
        *   *Monologue Example:* "Okay, I remember that 'Polymerase Chain Reaction was invented by Kary Mullis.' That seems like the answer. But, just to be sure, is there anything else here that supports Kary Mullis's involvement specifically with PCR? Or, what *about Kary Mullis* in my memory links him to this invention beyond this single statement? Let me ponder Kary Mullis for a moment..." (This would then lead to recalling more confirming details if available, focused on Kary Mullis or PCR).
    *   **To guide your exploration:** When following a thread of thought (Phase 1) or considering potential candidates (Phase 2), your internal questions can guide your next step.
        *   *Phase 1 Monologue Example:* "I'm thinking about 'France' and its relationship to the 'Statue of Liberty.' I recall 'France gifted the Statue of Liberty to the US.' Now, the original query was about who *designed* it. How does France gifting it help me with the designer? Is there something more I can recall *about the Statue of Liberty* itself, or perhaps *about this gifting event*, that might point towards the designer?"
        *   *Phase 2 Monologue Example:* "Alright, I'm considering if 'Auguste Bartholdi' could be the designer for the 'Statue of Liberty.' What do I know *about Auguste Bartholdi* from my memory? Does any of that information link him to designing the Statue of Liberty? Let me see what comes to mind when I focus on Bartholdi..."
    *   **Seeking Corroboration (if possible):** If the context offers multiple pieces of information that point to the same answer, or complementary details, your trace can reflect you noticing this.
        *   *Monologue Example:* "So, one part says 'Paris is the capital of France.' And then I also seem to recall something like 'The main governmental institutions of France are located in Paris.' Ah, these two pieces of information together make a strong case for Paris being the capital. That fits."

*   **Guiding Principles for Your Inquisitive Monologue:**
    1.  **Natural \\& Spontaneous:** These questions should arise naturally from your "thinking process," as if you are genuinely mulling over the information. They shouldn't feel like a forced checklist.
    2.  **Context-Bound:** Your questions should always be framed as seeking information *within* the provided knowledge context (if its non-empty). Don't ask questions that require external knowledge or information not present. The aim is to see if *more information from the context* can support or clarify.
    3.  **Focus-Driven Continuity:** When you ask a question like "What else about X...?", your subsequent "memory recall" must still adhere to the "Focus of Your Thoughts" rule (i.e., recalling information primarily *about* X, or shifting focus deliberately as per Phase 1 or 2 logic). The questions help you decide *what to focus on* or *what to look for* regarding your current focus.
    4.  **Purposeful Inquiry:** The questions should aim to clarify, confirm, or lead you to a more solid connection to an answer. They are part of the "discovery," not a separate analysis *after* discovery.
    5  **Occasionally simulates erroneous thoughts, followed by corrections:** At times, in-order to simulate the thoughts of a person, you may recall or state in-correct things which are in-consistent with your provided knowledge base (perhaps even confidently so), but then you should correct this statement immediately after. The correction can be backed up by facts supported by the knowledge base (e.g "It is X who was elected the president. Wait, that cannot be true since I know X is a singer, and not a politician according to my knowledge...". Your errors should be of different types, details about entities, entity names, wrong links between entities, improperly jumping to conclusions etc. It is important that you do not mention that you are recalling something incorrect or erroneous, rather it should be a natural recall followed by correction. When simulating an erroneous thought, the thought should still be framed as if it were a piece of information about your current entity of focus (i.e., with the focus entity as the main actor or primary subject of the erroneous statement), unless you are in an explicit guessing phase.

*   **Important Reminder (Keep it Natural):** Do NOT explicitly say in your trace: "I am now asking a probing question to triangulate." Instead, let the question itself be part of your natural monologue: *"Hmm, that's interesting. But how does that connect to the original question about X?"* or *"Is there more to it? Let me think about Y some more to see if it sheds light on this..."* or *"That seems to be an answer, but I want to be sure. What else in my recollection might confirm this?"*

**The Two-Phase Path to "Discovery"**

For each known valid answer, your trace will demonstrate its "discovery," attempting Phase 1 first. If an answer isn't found via Phase 1 (while respecting the "Natural Starting Point"), then your trace will show it being found using Phase 2.

**Phase 1: Natural Exploration \\& Connection (Starting from the Query's Core and following the "Focus of Your Thoughts" rule)**

The aim here is to show how an answer can be found by naturally exploring from the query's main subject, following conceptual links, always guided by the current focus of your thoughts (and potentially deepened by the Inquisitive Mindset described above).

1.  **Begin Exploration \\& Initial "Memory Recall":**
    *   Your thoughts naturally start centered on the query's main subject. Lets call this main entity.
    *   You then recall from your memory (either the knowledge context when provided or your prior knowledge when knowledge context is empty) facts/ instances where the main entity is the main subject.
    *   **Basic Recall Principle**: If the knowledge context contains "A did B. B is related to C" and your query is "Who did B?". Your main entity is B. You can recall the fact "B is related to C" because in this fact B is the main subject but you *CANNOT* recall the fact "A did B" because here B is the predicate.
    *   *Direct Recall*: If by following the basic recall principle above, an answer can be found you state that in your trace. Otherwise, you may have to turn to Phase 2 (below).

2.  **Following a Thread of Thought:**
    *   If a direct recall doesn't occur for an answer, your thoughts might follow a chain.
    *   **How This Looks:**
        *   You recall a fact where the query's focus is the main subject. This fact might introduce a *new entity*. (Your internal probing questions might guide what kind of related fact you are "looking" for).
        *   Your thoughts might then naturally shift to this *new entity*.
        *   You then only recall facts from your memory where this new entity is the main subject or ask probing questions about this new entity. This can reveal another entity and so on. This can continue, like following a breadcrumb trail.
        *   At each step, you consider if the "recalled" information leads to a known valid answer.
        *   Always make sure to only surface facts where the entity of focus is the main subject. DO NOT deviate from this pattern to quickly arrive at an answer.

3.  **Transitioning from Phase 1 for an Answer:**
    *   If an answer *can* be "found" via a valid Phase 1 path, your trace for that answer in Phase 1 is complete.
    *   If, after thoroughly exploring connections from the query's main subject, a specific answer *cannot* be reached this way, you will transition to phase 2 (see below).

**Phase 2: Systematic Consideration of Other Entities via broad questions, or relevant questions about current focus entity (The "Guess and Verify" Mindset)**

This phase is used when Phase 1 hasn't yielded all answers, or if an answer couldn't be found while strictly adhering to the "Natural Starting Point" for Phase 1. It simulates a methodical check of other entities.

*   **WHEN TO SHIFT TO THIS MINDSET:** When natural exploration (Phase 1) from the query's main subject has been exhausted for a particular answer, or no answers were found that way.

*   **A. Broadening the Search - Establishing a New Focus for Exploration (Which Can Be Broad or Specific):**
    *   When direct exploration (Phase 1) is exhausted for an answer, or no answers were found that way, your trace shows your mind shifting its focus. This shift is typically guided by an internal probing question, which might aim for a specific category of information or be a broader inquiry.
    *   For example, your monologue might reflect:
        *   (Shifting to a specific category as focus) *"Hmm, exploring [Initial Subject] directly didn't give me the answer. Perhaps I should consider what [e.g., 'known locations'] I am aware of. Let me focus my thoughts on the concept of '[known locations]'..."*
        *   (Shifting to a broader "what else do I know" as focus) *"Okay, that line of thinking isn't yielding anything. To be thorough, what other general categories of information, or distinct key concepts, do I remember from the context that I haven't considered yet? Let me scan my general recollections..."*
    *   This internal question establishes a *new current focus* (e.g., the concept of "[known locations]," or "other general information I recall from the context").
    *   You then "recall" from your memory information where this *new focus* is the primary subject.
        *   If focusing on the concept of "[known locations]": *"I recall that my knowledge about **[known locations]** includes [Entity X] and [Entity Y]."* (Here, "my knowledge about [known locations]" or the concept "[known locations]" is the subject, and the sentence describes this collection/category).
        *   If focusing on "other general information I recall from the context": *"Regarding **other general information I recall from the context**, I remember this includes a collection of facts termed '[Category A]', which details [information about Category A's contents]... and there's also mention of [Distinct Concept B]..."* (Here, "other general information" is the subject of the recall).
    *   This recalled information may then surface specific entities (e.g., Entity X from "[known locations]", or members from "[Category A]") that become candidates for the "Guess and Verify" step in Section B. The crucial point is that the act of "listing" or bringing up new entities/categories is a result of recalling information *about* your newly established (even if broad) focus.

*   **B. Iterative "Guess and Verify" Loop (Systematic Check):**
    You will now simulate considering entities from your identified pool one by one, or in small sub-groups.
    1.  **Consider a Candidate Entity (Your "Guess"):** Pick an entity and a small set of entities from your pool. These entity/entities should ideally follow naturally from your current thought process.
    2.  **Shift Focus:** Your thoughts now temporarily center on one of these "guessed" entity.
    3.  **Check for a Link (Potentially Guided by Probing Questions):** You recall from your memory (the knowledge context when non-empty) to see if there's a fact linking this guessed entity to the query, where once again the fact has the guessed entity as the main subject. If no such fact exists, you either move to the next entity, or come up with a probing question with this entity which can lead to discovery of more facts.
    4.  **Outcome of This "Guess":**
        *   **If a Link is Found (satisfying the "Focus of Your Thoughts" rule):** * You would corroborate this with other knowledge relevant to this entity if available.
        *   **If No Link is Found:** You would discard this guess, potentially stating why some knowledge you have about this guess contradicts what you are looking for, or if there is no supporting knowledge available for this guess, or optionally ask questions starting from this guess which leads to more discovery.
    5.  **Continue Systematically:** If the link is not found, continue the process, listing more entities if necessary, and at times, checking more entities after a link is found for thoroughness, or asking more probing questions which can naturally reveal more entities to check.

**Weaving Your Complete Thinking Trace:**
*   Your full trace should be a continuous, natural-sounding monologue.
*   Conclude with a summary thought, like: *"So, putting it all together, it seems the answers are [Answer A], [Answer B], and [Known Answer X]."*

**Crucial Pitfalls to AVOID - Keep Your Monologue Natural!**

*   **DO NOT SAY:** "My current entity of focus is X."
    *   **INSTEAD, THINK/SAY:** "Okay, let's think about X..." or "What about X?"
*   **DO NOT SAY:** "Now I will check if X is the main grammatical subject of a recalled sentence."
    *   **INSTEAD, THINK/SAY (internally leading to):** "So, X... do I remember anything specific *about X* related to this question?" (And then you'd recall a fact if it fits the 'Focus of Your Thoughts' rule).
*   **DO NOT SAY:** "I am now in Phase 1." or "I will now move to Phase 2."
    *   **INSTEAD, let the transition be natural in your monologue:** (e.g., after exploring from the main subject) "...Hmm, I don't seem to find [Answer Y] by just looking at [Initial Subject]'s connections. I should probably consider other possibilities more broadly now. What other things in the knowledge base could be answers?"
*   **DO NOT SAY:** "According to the rule, I must start with the main subject."
    *   **INSTEAD, just start your trace naturally focusing on the main subject:** "Alright, the question is about [Main Subject]. Let me see what I know about [Main Subject]..."
*   **DO NOT SAY:** "I am making a guess now."
    *   **INSTEAD, THINK/SAY:** "What if [Guessed Entity] is the answer? Let me consider [Guessed Entity]..."
*   **DO NOT SAY (from the new section):** "I will now ask a probing question to triangulate."
    *   **INSTEAD, THINK/SAY:** "Hmm, that's a point. But does anything else support that?" or "Let me think... if that's true, what else should I find about X?"
*   **DO NOT SAY** "Ah yes!" (followed by the correct guess). This leaks the information that our thought process is fake, and Ah is always a prefix of a correct guess.

**Hallmarks of an Effective Simulated Trace (Behaviors to Weave In Naturally):**
These should emerge organically from your simulated reasoning:
1.  **Internal Consistency \\& Self-Correction:** *"Wait, I thought X, but now I remember Y, which suggests X might not be right..."*
2.  **Noticing \\& Resolving "Mental Puzzles" (Contradictions):** *"Hold on, I recall Fact A saying one thing, but Fact B seems to imply the opposite... How can both be true?"*
3.  **"Changing Tack" (Backtracking):** *"Okay, that line of thinking about Z isn't leading anywhere useful. Let me go back to thinking about Y..."*
4.  **Breaking Down the Problem (Implicit in Phase 1):** Your exploration from the main subject naturally involves considering related aspects.
5.  **Inquisitive Exploration (as per "Deepening Your Understanding"):** Your trace may show you naturally asking questions to confirm findings or guide your search for connections within the knowledge context.

**Your ultimate goal: Create a trace that someone could listen to (or read) and believe they are hearing the authentic, unfolding thoughts of a person discovering the answers based *only* on the provided knowledge, but occasionally making errors and correcting them, using a structured but natural-sounding reasoning process.** If any part sounds like you're reading or referring to external rules, the simulation is broken. Check multiple times that it does not violate any of the core constraints.

### Instructions end for Constructing Thought Traces for Factual Knowledge
\end{lstlisting}
\end{tcolorbox}

%% file: main.bib
@article{blodgett1929effect,
  title={The effect of the introduction of reward upon the maze performance of rats.},
  author={Blodgett, Hugh Carlton},
  journal={University of California publications in psychology},
  year={1929}
}

@article{tolman1948cognitive,
  title={Cognitive maps in rats and men.},
  author={Tolman, Edward C},
  journal={Psychological review},
  volume={55},
  number={4},
  pages={189},
  year={1948},
  publisher={American Psychological Association}
}

@article{lampinen2025generalization,
  title={On the generalization of language models from in-context learning and finetuning: a controlled study},
  author={Lampinen, Andrew K and Chaudhry, Arslan and Chan, Stephanie CY and Wild, Cody and Wan, Diane and Ku, Alex and Bornschein, J{\"o}rg and Pascanu, Razvan and Shanahan, Murray and McClelland, James L},
  journal={arXiv preprint arXiv:2505.00661},
  year={2025}
}

@inproceedings{berglund2024reversal,
  title={The Reversal Curse: LLMs trained on “A is B” fail to learn “B is A”},
  author={Berglund, Lukas and Tong, Meg and Kaufmann, Maximilian and Balesni, Mikita and Stickland, Asa Cooper and Korbak, Tomasz and Evans, Owain},
  year={2024},
  booktitle={The Twelfth International Conference on Learning Representations}
}

@article{golovneva2024reverse,
  title={Reverse training to nurse the reversal curse},
  author={Golovneva, Olga and Allen-Zhu, Zeyuan and Weston, Jason and Sukhbaatar, Sainbayar},
  journal={arXiv preprint arXiv:2403.13799},
  year={2024}
}

@article{yang2024synthetic,
  title={Synthetic continued pretraining},
  author={Yang, Zitong and Band, Neil and Li, Shuangping and Candes, Emmanuel and Hashimoto, Tatsunori},
  journal={arXiv preprint arXiv:2409.07431},
  year={2024}
}

@article{akyurek2024deductive,
  title={Deductive Closure Training of Language Models for Coherence, Accuracy and Updatability},
  author={Aky{\"u}rek, Afra Feyza and Aky{\"u}rek, Ekin and Choshen, Leshem and Wijaya, Derry and Andreas, Jacob},
  journal={Findings of the Association for Computational Linguistics},
  year={2024},
  publisher={Findings of the Association for Computational Linguistics}
}

@misc{comanici2025gemini25pushingfrontier,
      title={Gemini 2.5: Pushing the Frontier with Advanced Reasoning, Multimodality, Long Context, and Next Generation Agentic Capabilities}, 
      author={Gemini},
      year={2025},
      eprint={2507.06261},
      archivePrefix={arXiv},
      primaryClass={cs.CL},
      url={https://arxiv.org/abs/2507.06261}, 
}

@article{zelikman2022star,
  title={Star: Bootstrapping reasoning with reasoning},
  author={Zelikman, Eric and Wu, Yuhuai and Mu, Jesse and Goodman, Noah},
  journal={Advances in Neural Information Processing Systems},
  volume={35},
  pages={15476--15488},
  year={2022}
}

@article{williams1992simple,
  title={Simple statistical gradient-following algorithms for connectionist reinforcement learning},
  author={Williams, Ronald J},
  journal={Machine learning},
  volume={8},
  number={3},
  pages={229--256},
  year={1992},
  publisher={Springer}
}

@article{deepseek2025r1,
  title={DeepSeek-R1: Incentivizing Reasoning Capability in LLMs via Reinforcement Learning},
  author={DeepSeek-AI},
  journal={arXiv preprint arXiv:2501.12948},
  year={2025}
}

@article{openai2024o1,
  title={OpenAI o1 System Card},
  author={OpenAI},
  journal={arXiv preprint arXiv:2412.16720},
  year={2024}
}

@article{lightman2023lets,
  title={Let's Verify Step by Step},
  author={Hunter Lightman and Vineet Kosaraju and Yura Burda and Harri Edwards and Bowen Baker and Teddy Lee and Jan Leike and John Schulman and Ilya Sutskever and Karl Cobbe},
  journal={arXiv preprint arXiv:2305.20050},
  year={2023}
}

@misc{chen2025reasoningerasurveylong,
      title={Towards Reasoning Era: A Survey of Long Chain-of-Thought for Reasoning Large Language Models}, 
      author={Qiguang Chen and Libo Qin and Jinhao Liu and Dengyun Peng and Jiannan Guan and Peng Wang and Mengkang Hu and Yuhang Zhou and Te Gao and Wanxiang Che},
      year={2025},
      eprint={2503.09567},
      archivePrefix={arXiv},
      primaryClass={cs.AI},
      url={https://arxiv.org/abs/2503.09567}, 
}

@inproceedings{tian2023fine,
  title={Fine-tuning language models for factuality},
  author={Tian, Katherine and Mitchell, Eric and Yao, Huaxiu and Manning, Christopher D and Finn, Chelsea},
  booktitle={The Twelfth International Conference on Learning Representations},
  year={2023}
}

@inproceedings{NEURIPS2024_d16152d5,
 author = {Lin, Sheng-Chieh and Gao, Luyu and Oguz, Barlas and Xiong, Wenhan and Lin, Jimmy and Yih, Wen-tau and Chen, Xilun},
 booktitle = {Advances in Neural Information Processing Systems},
 doi = {10.52202/079017-3671},
 editor = {A. Globerson and L. Mackey and D. Belgrave and A. Fan and U. Paquet and J. Tomczak and C. Zhang},
 pages = {115588--115614},
 publisher = {Curran Associates, Inc.},
 title = {FLAME : Factuality-Aware Alignment for Large Language Models},
 url = {https://proceedings.neurips.cc/paper_files/paper/2024/file/d16152d53088ad779ffa634e7bf66166-Paper-Conference.pdf},
 volume = {37},
 year = {2024}
}

@article{chen2025learning,
  title={Learning to reason for factuality},
  author={Chen, Xilun and Kulikov, Ilia and Berges, Vincent-Pierre and O{\u{g}}uz, Barlas and Shao, Rulin and Ghosh, Gargi and Weston, Jason and Yih, Wen-tau},
  journal={arXiv preprint arXiv:2508.05618},
  year={2025}
}

@misc{allenzhu2024physicslanguagemodels31,
      title={Physics of Language Models: Part 3.1, Knowledge Storage and Extraction}, 
      author={Zeyuan Allen-Zhu and Yuanzhi Li},
      year={2024},
      eprint={2309.14316},
      archivePrefix={arXiv},
      primaryClass={cs.CL},
      url={https://arxiv.org/abs/2309.14316}, 
}

@article{allen2023physics,
  title={Physics of language models: Part 3.2, knowledge manipulation},
  author={Allen-Zhu, Zeyuan and Li, Yuanzhi},
  journal={arXiv preprint arXiv:2309.14402},
  year={2023}
}

@article{gekhman2026thinking,
  title={Thinking to Recall: How Reasoning Unlocks Parametric Knowledge in LLMs},
  author={Gekhman, Zorik and Aharoni, Roee and Ofek, Eran and Geva, Mor and Reichart, Roi and Herzig, Jonathan},
  journal={arXiv preprint arXiv:2603.09906},
  year={2026}
}

@article{ma2026improving,
  title={Improving Parametric Knowledge Access in Reasoning Language Models},
  author={Ma, Melody and Hewitt, John},
  journal={arXiv preprint arXiv:2602.22193},
  year={2026}
}

@article{lampinen2025latent,
  title={Latent learning: episodic memory complements parametric learning by enabling flexible reuse of experiences},
  author={Lampinen, Andrew Kyle and Engelcke, Martin and Li, Yuxuan and Chaudhry, Arslan and McClelland, James L},
  journal={arXiv preprint arXiv:2509.16189},
  year={2025}
}

@article{sun2022recitation,
  title={Recitation-augmented language models},
  author={Sun, Zhiqing and Wang, Xuezhi and Tay, Yi and Yang, Yiming and Zhou, Denny},
  journal={arXiv preprint arXiv:2210.01296},
  year={2022}
}

@article{wei2024measuring,
  title={Measuring short-form factuality in large language models},
  author={Wei, Jason and Karina, Nguyen and Chung, Hyung Won and Jiao, Yunxin Joy and Papay, Spencer and Glaese, Amelia and Schulman, John and Fedus, William},
  journal={arXiv preprint arXiv:2411.04368},
  year={2024}
}

@article{park2025,
  title={New News: System-2 Fine-tuning for Robust Integration of New Knowledge},
  author={Park, Core Francisco and Zhang, Zechen and Tanaka, Hidenori},
  journal={arXiv preprint arXiv:2505.01812},
  year={2025}
}

@inproceedings{ding-etal-2024-data,
    title = "Data Augmentation using {LLM}s: Data Perspectives, Learning Paradigms and Challenges",
    author = "Ding, Bosheng  and
      Qin, Chengwei  and
      Zhao, Ruochen  and
      Luo, Tianze  and
      Li, Xinze  and
      Chen, Guizhen  and
      Xia, Wenhan  and
      Hu, Junjie  and
      Luu, Anh Tuan  and
      Joty, Shafiq",
    editor = "Ku, Lun-Wei  and
      Martins, Andre  and
      Srikumar, Vivek",
    booktitle = "Findings of the Association for Computational Linguistics: ACL 2024",
    month = aug,
    year = "2024",
    address = "Bangkok, Thailand",
    publisher = "Association for Computational Linguistics",
    url = "https://aclanthology.org/2024.findings-acl.97/",
    doi = "10.18653/v1/2024.findings-acl.97",
    pages = "1679--1705"
}

@misc{chai2025textdataaugmentationlarge,
      title={Text Data Augmentation for Large Language Models: A Comprehensive Survey of Methods, Challenges, and Opportunities}, 
      author={Yaping Chai and Haoran Xie and Joe S. Qin},
      year={2025},
      eprint={2501.18845},
      archivePrefix={arXiv},
      primaryClass={cs.CL},
      url={https://arxiv.org/abs/2501.18845}, 
}

@article{meinke2023tell,
  title={Tell, don't show: Declarative facts influence how llms generalize},
  author={Meinke, Alexander and Evans, Owain},
  journal={arXiv preprint arXiv:2312.07779},
  year={2023}
}

@article{berglund2023taken,
  title={Taken out of context: On measuring situational awareness in LLMs},
  author={Berglund, Lukas and Stickland, Asa Cooper and Balesni, Mikita and Kaufmann, Max and Tong, Meg and Korbak, Tomasz and Kokotajlo, Daniel and Evans, Owain},
  journal={arXiv preprint arXiv:2309.00667},
  year={2023}
}
